\documentclass[runningheads]{llncs}
\usepackage[T1]{fontenc}
\usepackage{graphicx}
\usepackage{hyperref}
\usepackage{amsmath}
\usepackage{amsfonts}
\usepackage{xcolor}
\usepackage{pgfplots}
\usepackage{wrapfig}
\usepackage{graphicx} 
\usepackage{adjustbox}
\usepackage{amsfonts}
\usepackage{amsmath,amssymb}
\usepackage{mdframed}

\usepackage{todonotes}

\usepackage{tikz}
\usetikzlibrary{shapes.geometric, arrows, positioning}
\usetikzlibrary{fit}  
\usetikzlibrary{backgrounds}
\usepackage[ruled,vlined]{algorithm2e}

\usepackage{subcaption}  

\usepackage{float}

\newcommand{\C}[1]{\mathcal{#1}}
\newcommand{\B}[1]{\mathbb{#1}}

\newcounter{myctr}

\begin{document}
\title{Locally Pareto-Optimal Interpretations for\\ Black-Box Machine Learning Models\thanks{The work of S. Chakraborty was partly supported by a Qualcomm Faculty Award. The work of S. Akshay was partly supported by the SBI Foundation Hub for Data Science \& Analytics, IIT Bombay. The UC Berkeley authors were partly supported by the DARPA Provably Correct Design of Adaptive Hybrid Neuro-Symbolic Cyber Physical Systems (ANSR) program award number FA8750-23-C-0080, and by Nissan and Toyota under the iCyPhy Center. The work of H. Torfah was partly supported by the Wallenberg AI, and Autonomous Systems and Software Program (WASP) funded by the Knut and Alice Wallenberg Foundation. \\ This work has been accepted at ATVA 2025.}}

\titlerunning{Locally PO interpretations for Black-Box ML models}

\author{Aniruddha Joshi\inst{1} \and Supratik	Chakraborty\inst{2} \and S Akshay\inst{2} \and \\ Shetal Shah\inst{2} \and Hazem Torfah\inst{3} \and Sanjit Seshia\inst{1}}
\institute{University of California at Berkeley, USA 
\\ \email{\{aniruddhajoshi, sseshia\}@berkeley.edu} \and Indian Institute of Technology Bombay, India
\\ \email{\{ supratik, akshayss, shetals\}@cse.iitb.ac.in} \and Chalmers University of Technology and University of Gothenburg. Sweden
\\ \email{hazemto@chalmers.se}}

%
%
 \authorrunning{A. Joshi et al.}


\maketitle              
%

\begin{abstract} Creating meaningful interpretations for black-box machine learning models involves balancing two often conflicting objectives: accuracy and explainability. Exploring the trade-off between these objectives is essential for developing trustworthy interpretations. While many  techniques for multi-objective interpretation synthesis have been developed, they typically lack formal guarantees on the Pareto-optimality of the results. Methods that do provide such guarantees, on the other hand, often face severe scalability limitations when exploring the Pareto-optimal space.
To address this, we develop a framework based on local optimality guarantees that enables more scalable synthesis of interpretations. Specifically, we consider the problem of synthesizing a set of Pareto-optimal interpretations with local optimality guarantees, within the immediate neighborhood of each solution. Our approach begins with a multi-objective learning or search technique, such as Multi-Objective Monte Carlo Tree Search, to generate a best-effort set of Pareto-optimal candidates with respect to accuracy and explainability. We then verify local optimality for each candidate as a Boolean satisfiability problem, which we solve using a SAT solver.
We demonstrate the efficacy of our approach on a set of  benchmarks, comparing it against previous methods for exploring the Pareto-optimal front of interpretations.  In particular, we show that our approach yields interpretations that closely match those synthesized by methods offering global guarantees.

\end{abstract}

\section{Introduction}
The use of machine learning (ML) components is rapidly increasing across various applications and domains. However, the complexity of ML models often makes it difficult to understand how these components function. As a result, their behavior is frequently treated as a black box. To build trust in ML models, especially in domains where accountability and safety are critical, it is essential to provide interpretations that accurately reflect the underlying functionality of the ML model, while being also understandable by humans.


Over the past decade, a vast body of research has focused on explaining the behavior of ML models (e.g, see \cite{xaisurvey} for a survey). A prominent approach in this area has been the development of post-hoc interpretation methods, i.e., techniques used to explain the behavior of an ML model after it has been trained. These methods often involve generating surrogate models (e.g., decision trees, decision lists, etc.) that serve as simplified approximations of the original complex model \cite{xaisurvey}. A key challenge in creating such models lies in finding the right balance between accuracy and explainability. In many cases, these two objectives are in direct conflict: a simple, human-understandable explanation may diverge significantly from the predictions of the original model, while a more accurate surrogate may be too complex for meaningful human interpretation. This  raises a central question: how much accuracy are users willing to sacrifice for the sake of explainability, and vice versa?

While there are no easy answers to the above problem, a promising way to address it is by exploring the Pareto-optimal space of interpretations \cite{DBLP:conf/fmcad/TorfahSCAS21}. Rather than generating a single interpretation optimized for one objective (or even for a weighted combination of objectives), the Pareto-optimal interpretation problem involves examining the trade-off across multiple objectives, such as the balance between accuracy and explainability. This approach can be very beneficial for users aiming to understand these trade-offs; however, it also introduces significant challenges, particularly in efficiently navigating the space of possible interpretations. Previous work has demonstrated that it is possible to explore this space with formal statistical guarantees on the explored Pareto-optimal curve \cite{DBLP:conf/fmcad/TorfahSCAS21}. However, these approaches often struggle to scale due to their full reliance on computationally intensive symbolic techniques like constraint solvers. In this paper, we address the question of whether lighter and faster methods can be used to generate candidate Pareto-optimal interpretations, which can then be refined using more powerful tools such as Boolean satisfiability solvers.

Specifically, we propose the use of Multi-Objective Monte Carlo Tree Search (MO-MCTS)~\cite{DBLP:journals/jmlr/WangS12}, a well-established technique from the field of reinforcement learning known for its strong empirical performance, to quickly approximate the Pareto-optimal curve. We integrate MO-MCTS with Boolean satisfiability solving, enabling us to obtain local guarantees on the Pareto-optimality of interpretations. This replaces  earlier global Pareto-optimality guarantees with theoretically weaker local Pareto-optimality. However, since MO-MCTS is capable of recovering global guarantees over time, running our approach for a sufficient number of iterations yields interpretations whose local guarantees converge toward global ones. 
In our experiments, we show empirical evidence that the local guarantees obtained using MO-MCTS within a reasonable timeout are indeed close to the global guarantees.  The notion of locally Pareto-optimal interpretations, that we introduce, also offers additional benefits. Theoretically, they support anytime guarantees, meaning that useful solutions can be extracted at any point during execution (unlike earlier MaxSAT-based approaches). Additionally, they capture locally imperturbable interpretations, where small changes in the interpretation do not alter the optimality within a local neighborhood.
%

In summary our contributions are the following:
\begin{enumerate}
    \item We formalize the problem of locally Pareto-optimal interpretation synthesis for black-box ML models, where explainability is traded off for accuracy under user control.
    \item We develop a new technique to solve the above problem by a two phase hybrid algorithm that integrates Multi-objective Monte-Carlo Tree search with Boolean satisfiability solving. 
    \item We show that our approach converges monotonically to the global optimal, with local optimality guarantees for early stopping. 
    \item Our experimental results show that in practice, our approach obtains results close to the global optimal in several benchmarks. For larger benchmarks, we obtain results (with local optimality guarantees), where earlier approaches fail to produce \emph{any} globally Pareto-optimal solution.
\end{enumerate}

\paragraph{Related Work}
The literature contains a significant body of work focused on methods for interpreting black-box ML models.  In certain applications, the goal is to explain the output of a black-box model in the vicinity of a specific input. To achieve this, specialized techniques have been developed that provide local and robust explanations \cite{DBLP:journals/corr/abs-1805-10820,DBLP:conf/nips/LundbergL17,DBLP:conf/naacl/Ribeiro0G16,DBLP:conf/aaai/Ribeiro0G18}.
Other
applications use techniques like generation of surrogate models and model distillation (in the form of decision trees \cite{DBLP:journals/pr/KrishnanSB99,DBLP:conf/nips/CravenS95,DBLP:conf/cec/JohanssonKLB13,DBLP:conf/sigmod/Krishnan017,DBLP:conf/cp/YuISB20}). For further information on these techniques,
we refer the reader to the surveys in \cite{DBLP:journals/access/AdadiB18,xaisurvey}.

The problem of synthesizing multi-objective interpretations of black-box ML models was introduced and formalized in \cite{DBLP:conf/fmcad/TorfahSCAS21}. In that work, a MaxSAT-based engine was used to synthesize a set of Pareto-optimal interpretations. This approach provides statistical global guarantees on the accuracy of the synthesized interpretations.
Similar ideas were explored in \cite{DBLP:conf/cp/YuISB20}, where the authors encoded the problem of finding an interpretation as optimal decision sets, and in \cite{DBLP:conf/aaai/ZhangXSR23}, where sparse optimal decision trees were constructed using an objective function combining misclassification rate and the number of leaves. In comparison to the approach in \cite{DBLP:conf/fmcad/TorfahSCAS21}, the solutions presented in \cite{DBLP:conf/cp/YuISB20} and \cite{DBLP:conf/aaai/ZhangXSR23} provide a single point in the Pareto-optimal space, yielding a single value for correctness and explainability measures.
Our approach aligns with the goals of \cite{DBLP:conf/fmcad/TorfahSCAS21} in providing Pareto-optimal interpretations with guarantees on correctness. 
However, in contrast to \cite{DBLP:conf/fmcad/TorfahSCAS21}, our method offers a much more scalable solution for exploring the Pareto-optimal front.  In \cite{surrogate2025}, the authors use multi objective optimization to find locally optimal Pareto solutions to  the joint problem of generating a machine learning model and its surrogate. Our work is agnostic to the machine learning model, which is assumed to be a black box. 

Monte-Carlo Tree Search (MCTS) is a highly effective procedure for heuristically searching through a complex space of actions and rewards. Browne et al.~\cite{DBLP:mcts_survey} provide an excellent review on MCTS techniques and heuristics, and Swiechowski et al.~\cite{DBLP:mcts_new_survey} give a review of recent modifications and advances on domain specific adaptations required for MCTS. Other papers~\cite{DBLP:journals/jmlr/WangS12,mo-mcts2} consider the multi-objective variant of MCTS that is relevant to our work.



\section{Preliminaries and Problem Formulation}\label{sec:problemformulation}

We borrow some terminology from~\cite{DBLP:conf/fmcad/TorfahSCAS21} in presenting the problem formulation below.
Let $\C{I}$ be an input domain, $\C{O}$ be an output domain, and $\Delta(\C{I}\times\C{O})$ be a distribution over the input-output space induced by a black-box ML model.
Let $\C{G}$ denote the class of interpretations mapping $\C{I}$ to $\C{O}$. 
For each interpretation $g\in \C{G}$, we abuse notation and define the semantic function $g:\C{I}\rightarrow\C{O}$ that outputs $g(i)\in\C{O}$ for every input $i\in\C{I}$.
We also assume that each interpretation $g\in\C{G}$ is associated with a non-negative real-valued \emph{correctness measure} $\C{C}(g)\in \B{R}_{\geq 0}$, and a non-negative real-valued \emph{explainability measure} $\C{E}(g) \in \B{R}_{\geq 0}$.
Intuitively, $\C{C}(g)$ measures the accuracy of the interpretation $g$ w.r.t. the input-output distribution $\Delta(\C{I}\times \C{O})$, and {\color{black} $\C{E}(g)$ is a user-specified measure of the human understandability of the interpretation.} 
 In general we want to synthesize interpretations with high values of correctness and explainability measures.

For an interpretation $g \in \C{G}$, we define $(\C{C}(g), \C{E}(g))$ to be the \emph{goodness tuple} of $g$.  The partial order $\preceq$ on goodness tuples is defined as follows: $(c,e)\preceq (c',e')$ iff $c\leq c'$ and $e\leq e'$. The strict partial-order $\prec$ on such tuples is defined as: 
$(c,e)\prec(c',e')$ iff $(c,e)\preceq (c',e')$ and at least one of $c<c'$ or $e<e'$ holds. Given a set $\C{S}$ of goodness measures $(c,e)$, we define $\max^{\preceq} ~\C{S}$ to be the set of all $\preceq$-maximal tuples in $\C{S}$.

\begin{definition}[Pareto-Domination and Optimality~\cite{Luc2008,DBLP:conf/fmcad/TorfahSCAS21}]
Given a class $\C{G}$ of interpretations mapping $\C{I}$
to $\C{O}$, we say an interpretation $g$ Pareto-dominates another interpretation $g'$ if $(\C{C}(g'), \C{E}(g')) \prec (\C{C}(g), \C{E}(g))$.  An interpretation $g$ is said to be \textbf{Pareto-optimal} (or PO) in $\C{G}$ if $g \in \C{G}$, and there is no other interpretation $g' \in \C{G}$ that Pareto-dominates $g$.  In other words, $(\C{C}(g), \C{E}(g))$ is $\preceq$-maximal in the set of goodness tuples of all interpretations in $\C{G}$.  
\end{definition}




In general, there may be multiple PO interpretations
in $\C{G}$, with one or more interpretations corresponding to each $\preceq$-maximal tuple $(c, e)$.  
In~\cite{DBLP:conf/fmcad/TorfahSCAS21}, Torfah et al presented a technique to synthesize a set of PO interpretations, one for every $\preceq$-maximal tuple $(c, e)$.  This effectively gives the user a set of interpretations, none of which can be "improved" on both the accuracy and explainability measures together.  Unfortunately, computing such PO interpretations is computationally expensive, and the technique of~\cite{DBLP:conf/fmcad/TorfahSCAS21} does not scale large to problem instances.  This motivates us to define
locally Pareto-optimal interpretations.

\begin{definition}[Locally Pareto-optimal interpretation]\label{def:localpoi}
    Let $\C{G}$ be a class of interpretations mapping $\C{I}$ to $\C{O}$, with correctness measure $\C{C}(\cdot)$ and explainability
    measure $\C{E}(\cdot)$, as before.  Let $\delta_c \in \B{R}_{\geq 0}$ be the \emph{correctness slack}, and $\delta_e\in\B{R}_{\geq 0}$ be the \emph{explainabiility slack}, where $\B{R}_{\geq 0}$ denotes the set of all
    non-negative reals.  
    We say that an interpretation $g\in\C{G}$ is locally Pareto-optimal (LPO) w.r.t $\delta_c$ and $\delta_e$ if $\nexists g' \in \C{G}$ such that $(\C{C}(g),\C{E}(g))\prec (\C{C}(g'),\C{E}(g'))\preceq (\C{C}(g)+\delta_c, \C{E}(g)+\delta_e)$.  
\end{definition}
Informally, if interpretation $g$ is LPO w.r.t. $\delta_c$ and $\delta_e$, then there is no other interpretation $g'$ that Pareto-dominates $g$, while having its correctness (respectively, explainability) measure within a window of $\delta_c$ (respectively, $\delta_e$) of the corresponding measure for $g$.  Note that an LPO interpretation $g$ may not be a PO interpretation.  However, there cannot be a "nearby" (in the correctness-explainability measure space) interpretation $g'$ that improves the correctness (respectively, explainability) measure of $g$ by an amount bounded by $\delta_c$ (respectively, $\delta_e$) without adversely affecting explainability (respectively, correctness).  Clearly, every PO intepretation is also LPO w.r.t $\delta_c$ and $\delta_e$, for every $\delta_c, \delta_e \in \B{R}_{\geq 0}$.  We can now define the problem of synthesizing LPO interpretations as follows.

\begin{mdframed}{
\begin{problem} [Locally Pareto-Optimal Interpretation Synthesis] 
        Let $\C{G}$, $\C{C}(\cdot)$, $\C{E}(\cdot), \delta_c$ and $\delta_e$ be as in Definition~\ref{def:localpoi}. 
        The \emph{locally Pareto-optimal (or LPO)} interpretation synthesis problem requires us to synthesize a set $\C{S}$ of interpretations in $\C{G}$ s.t. every $g \in \C{S}$ is LPO w.r.t $\delta_c$ and $\delta_e$, and every distinct $g, g' \in \C{S}$ are incomparable, i.e. neither of them Pareto-dominates the other. 
    \label{def:approx_poisyn}
\end{problem}
}
\end{mdframed}

A few points about Problem~\ref{def:approx_poisyn} are worth noting.  First, the requirement that interpretations in the solution must not Pareto-dominate each other ensures that we don't report LPO interpretations, one of which dominates the other.  This can indeed happen for LPO interpretations $g$ and $g'$ if, for example, $\C{C}(g) > \C{C}(g') + \delta_c$ and $\C{E}(g) > \C{E}(g') + \delta_e$.  In such cases,  $g$ is a "better" interpretation than $g'$, and it is not meaningful to report both $g$ and $g'$.  Second, every set of PO interpretations serves as a solution to Problem~\ref{def:approx_poisyn}, for every non-negative value of $\delta_c$ and $\delta_e$.  However, not every LPO interpretation may be a PO interpretation in general.  Hence, the ask of Problem~\ref{def:approx_poisyn} is weaker than that of finding PO interpretations (as addressed in~\cite{DBLP:conf/fmcad/TorfahSCAS21}). Third, if we set $\delta_c$ and $\delta_e$ to $M$ and $N$ respectively, where $M > \max_{g \in \C{G}} \C{C}(g)$ and $N > \max_{g \in \C{G}} \C{E}(g)$, then every solution to Problem~\ref{def:approx_poisyn} also gives a set of PO interpretations.  Hence, Problem~\ref{def:approx_poisyn} can be technically used to obtain PO interpretations, if we know $\max_{g \in \C{G}} \C{C}(g)$ and $\max_{g \in \C{G}} \C{E}(g)$.  Finally, note that Problem~\ref{def:approx_poisyn} does not require us to synthesize a maximal (by subset ordering) set of incomparable LPO interpretations. This makes it possible to synthesize increasingly larger (set cardinality-wise) solutions incrementally. 
\section{Our Approach and the Problem Instantiation}
In order to solve Problem~\ref{def:approx_poisyn}, at a high level, we use a hybrid approach that has two phases.
In the first phase, we use techniques developed from the multi-objective learning/search literature. 
We assume a time budget for the first phase, and restrict ourselves to those techniques that maintain a best-effort solution at all times. 
Once the time budget is exhausted, we stop the first phase, retrieve the best-effort interpretations from the first phase, and then pass these interpretations to the second phase.
We assume no guarantees about finding (locally) Pareto-optimal solutions from the first phase, since most multi-objective learning/search based techniques do not give strong anytime theoretical guarantees. 

\begin{figure}
    \begin{minipage}{0.45\textwidth}
    \includegraphics[width=0.96\textwidth]{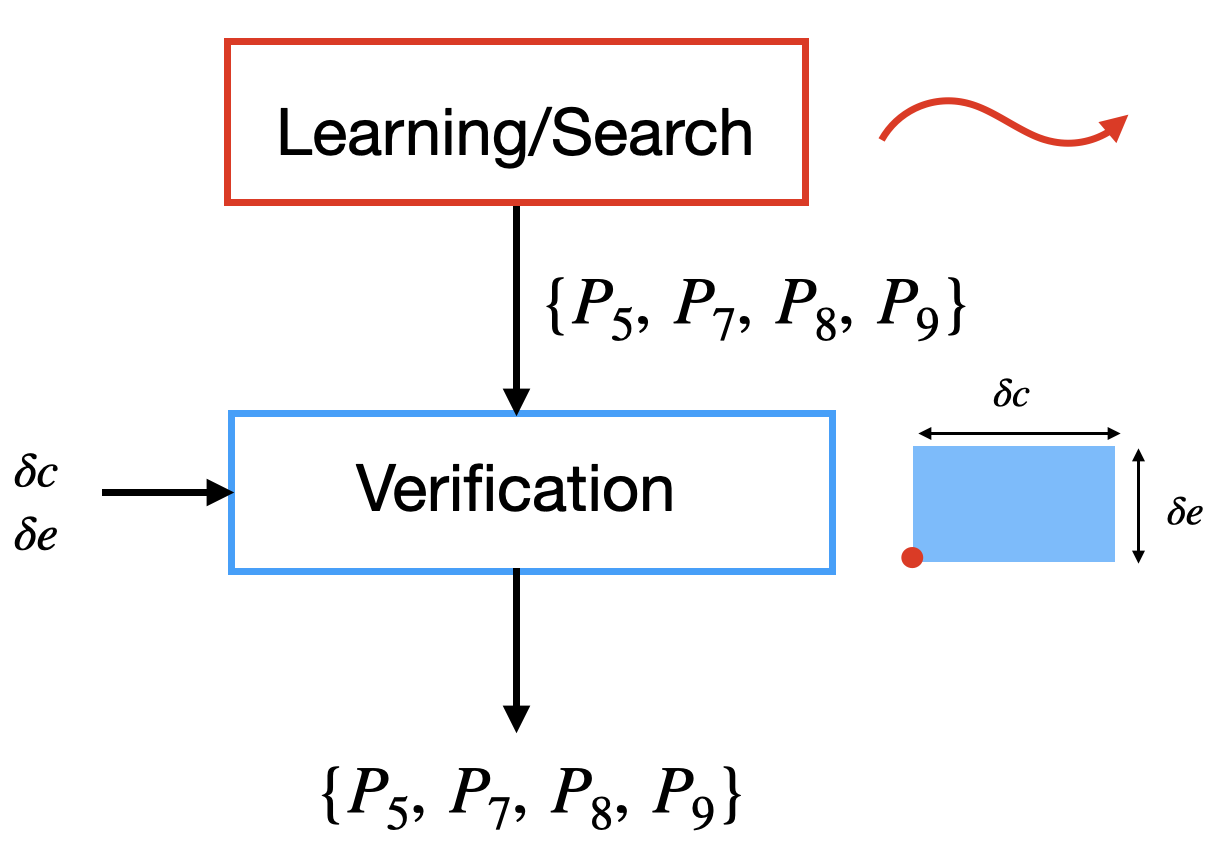}
    \subcaption{The red curvy arrow represents a multi-objective search procedure with a time budget. If this outputs interpretations $\{P_5, P_7, P_8, P_9\}$, we then verify absence of a dominating solution within slack $\delta_c$ and $\delta_e$, as represented by the blue rectangle.}
    \label{fig:oap}

    \end{minipage}
    \hspace{0.3cm}
    \begin{minipage}{0.45\textwidth}
    \includegraphics[width=0.8\textwidth]{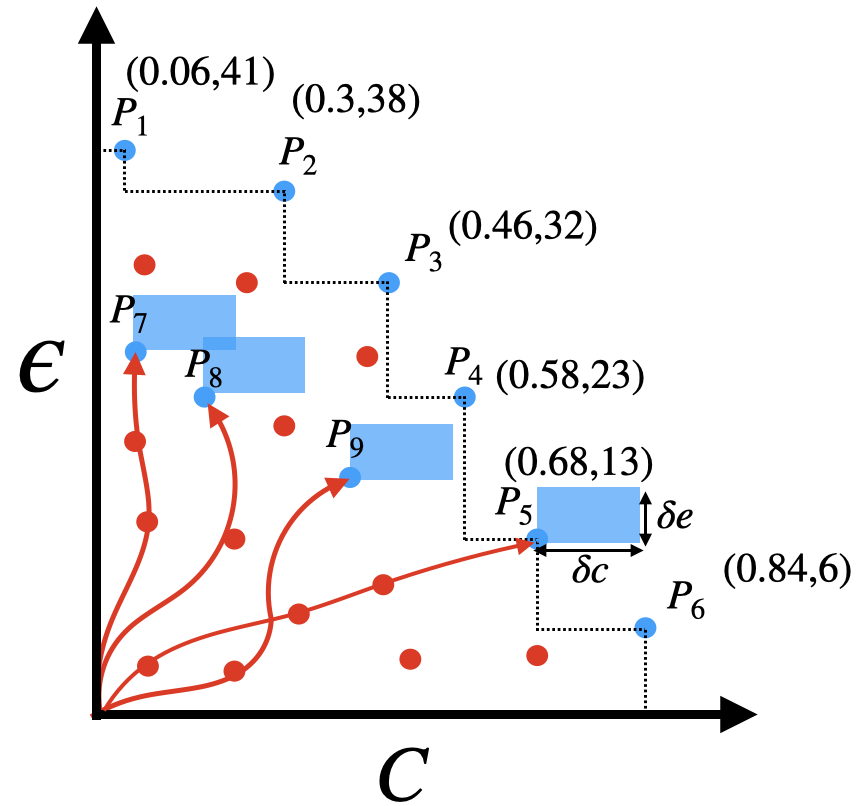}
    \subcaption{Illustrating our navigation of the search space. The x-axis (resp. y-axis) represents the correctness (resp. explainability) measure. The red curvy arrows indicate the search based phase, and the blue rectangles indicate the verification phase.  $P_1, \ldots P_6$ are PO interpretations. Our approach finds the LPO interpretations $P_5, P_7, P_8, P_9$}
    \label{fig:oa}
     \end{minipage}
     \caption{Our approach}
    \label{fig:oas}

\end{figure}

In the second phase, we take as inputs the set of interpretations obtained from the first phase, and also the correctness and explainability slacks $\delta_c$ and $\delta_e$.  We then
verify that the interpretations obtained from the first phase are incomparable and also locally Pareto-optimal w.r.t. the given slacks.  
We do this in two steps. In the first step, we verify that the interpretations are incomparable.  If they are not, we discard dominated interpretations.
Then in the second step, we verify that the remaining interpretations are locally Pareto-optimal.
If any of the interpretations are not locally Pareto-optimal, then we improve upon those interpretations.
Thus, at the end we are able to output those interpretations that are incomparable and also locally Pareto-optimal. 
%
Figure~\ref{fig:oas} diagrammatically shows the two phases in our approach.

\subsection{Instantiating the Interpretations and Measures}
\label{sec:fixing}
In order to solve Problem~\ref{def:approx_poisyn}, we must first choose the class $\C{G}$ of interpretations, and the explainability and correctness measures $\C{C}(\cdot)$ and $\C{E}(\cdot)$.
%
We start by choosing decision trees as the class of interpretations, since these are  the most widely used  in the explainable AI literature. 
Note that the work of~\cite{DBLP:conf/fmcad/TorfahSCAS21} considers decision diagrams for interpretations.
Decision trees are essentially decision diagrams with the additional restriction that every node has at most one incoming edge.  Hence, we can use the same correctness and
explainability measures as used
in~\cite{DBLP:conf/fmcad/TorfahSCAS21} , making it easier to compare our results with those of~\cite{DBLP:conf/fmcad/TorfahSCAS21}.

\subsubsection{Class of Interpretations: Decision Trees}

\begin{figure}
    \begin{minipage}{0.45\textwidth}
\begin{tikzpicture}[every node/.style={font=\footnotesize}, scale=0.9, transform shape]
    \tikzstyle{decision} = [draw, diamond, aspect=2, rounded corners=0.8ex, text centered, inner sep=1pt, minimum width=1.5cm, minimum height=1cm]

    \tikzstyle{process} = [rectangle, draw, text centered, rounded corners]
    \tikzstyle{startstop} = [rectangle, rounded corners, draw, fill=gray!20]
    \tikzstyle{alert} = [rectangle, rounded corners, draw, fill=red!20]
    
    \node (clouds) [decision] {\textit{clouds}};
    \node (time1) [decision, below left=1cm and 1cm of clouds] {\textit{time}};
    \node (alert0) [alert, below=1cm of clouds] {\textbf{alert}};
    \node (time2) [decision, below right=1cm and 1cm of clouds] {\textit{time}};

    \draw[->] (clouds) -- node[midway, left] {2} (time1);
    \draw[->] (clouds) -- node[midway, left] {1} (alert0);
    \draw[->] (clouds) -- node[midway, right] {3} (time2);

    \node (alert1) [alert, below right=1cm and 0.3cm of time1] {\textbf{alert}};
    \node (pos) [decision, below left=1cm and 0.3cm of time1] {\textit{pos}};

    \draw[->] (time1) -- node[pos=0.4, right] {2} (alert1);
    \draw[->] (time1) -- node[pos=0.4, left] {1} (pos);

    \node (alert2) [alert, below right=1cm and 0.3cm of time2] {\textbf{alert}};
    \node (noalert) [startstop, below left=1cm and -0.5cm of time2] {\textbf{no alert}};

    \draw[->] (time2) -- node[pos=0.4, right] {1} (alert2);
    \draw[->] (time2) -- node[pos=0.4, left] {2} (noalert);
    
    \node (alert2) [alert, below left=1cm and 0.3cm of pos] {\textbf{alert}};
    \node (noalert) [startstop, below right=1cm and -0.5cm of pos] {\textbf{no alert}};

    \draw[->] (pos) -- node[pos=0.4, left] {1} (alert2);
    \draw[->] (pos) -- node[pos=0.4, right] {2} (noalert);
\end{tikzpicture}     
        \subcaption{Decision Tree with functions $\{clouds$, $time$, $pos\}$ and labels $\{alert$, $no$ $alert\}$. The functions $time$ and $pos$ have two outgoing branches, and $clouds$ has tree.}
        \label{fig:dt}

    \end{minipage}
    \hspace{2cm}
    \begin{minipage}{0.4\textwidth}
\begin{tikzpicture}[every node/.style={font=\footnotesize}, scale=0.9, transform shape]
    \tikzstyle{decision} = [draw, diamond, aspect=2, rounded corners=0.8ex, text centered, inner sep=1pt, minimum width=1.5cm, minimum height=1cm]

    \tikzstyle{process} = [rectangle, draw, text centered, rounded corners,fill=teal!20]
    \tikzstyle{startstop} = [rectangle, rounded corners, draw, fill=gray!20]
    \tikzstyle{alert} = [rectangle, rounded corners, draw, fill=red!20]
    
    \node (clouds) [decision] {\textit{clouds}};
    \node (time1) [decision, below left=1cm and 1cm of clouds] {\textit{time}};
    \node (alert0) [alert, below=1cm of clouds] {\textbf{alert}};
    \node (N2) [process, below right=0.7cm and 0.7cm of clouds] {\textit{N}};

    \draw[->] (clouds) -- node[midway, left] {2} (time1);
    \draw[->] (clouds) -- node[midway, left] {1} (alert0);
    \draw[->] (clouds) -- node[midway, right] {3} (N2);

    \node (alert1) [alert, below right=1cm and 0.3cm of time1] {\textbf{alert}};
    \node (N1) [process, below left=1cm and 0.3cm of time1] {\textit{N}};

    \draw[->] (time1) -- node[pos=0.4, right] {2} (alert1);
    \draw[->] (time1) -- node[pos=0.4, right] {1} (N1);


\end{tikzpicture}
        \subcaption{Partial decision tree with two unassigned nodes labeled $N$ be default.}
        \label{fig:PDT}
     \end{minipage}
     \caption{(a) Complete and (b) Partial Decision Tree}
    \label{fig:DTs}

\end{figure}
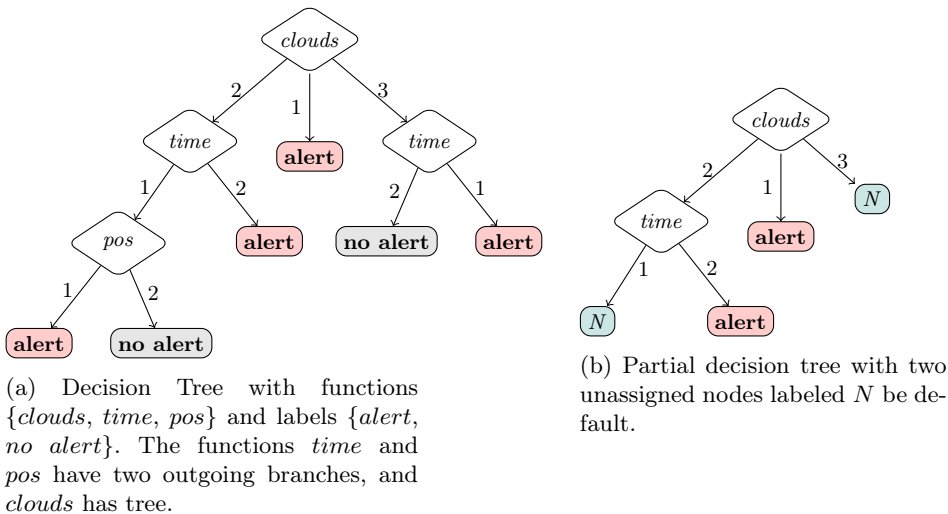

Let $\mathcal I$ and $\mathcal O$ denote the input and output domains, respectively. 
We restrict the output domain $\C{O}$ to a finite set $L:=\{l_1, l_2, \ldots l_{|L|}\}$ of labels. 
Let $F:=\{f_1, f_2, \ldots, f_{|F|}~|~f_i:\mathcal{I}\rightarrow \{1,2,\ldots b_i\}\}$ be a set of functions $f_i$ that map the input domain $\mathcal{I}$ to a finite set of branches $\{1,2,\ldots b_i\}$. 
We use $L$ and $F$ to label the nodes of decision trees. 
Specifically, each internal node of a tree has an incoming edge and outgoing edges, and is labeled by a function $f_i$. 
Each branch $j\in\{1,2,\ldots b_i\}$ in the range of $f_i$ corresponds to an outgoing edge of the internal node. 
The leaf nodes in a decision tree are labeled by $L$.
Figure~\ref{fig:dt} gives an example of a decision tree.

In order to represent (possibly partially constructed) decision trees at intermediate steps of our multi-objective search procedure, we consider a context-free grammar~\cite{DBLP:books/aw/HopcroftU79} $\mathfrak{G}$ that allows us to generate decision trees in a principled manner.  
Grammar $\mathfrak{G}$ uses the symbol $N$ as the starting (non-terminal) symbol, and $\Sigma:=\{$ $[$, $]$, $\raisebox{0.2em}{,}$~, $f_1$, $f_2$, $\ldots f_{|F|}$, $l_1$, $l_2$, $\ldots l_{|L|}\}$ as the set of terminal symbols.  It has the following production rules:
$$N:=f_1[\underbrace{N, N, \ldots N}_{b_1~\text{times}}]~|~f_2[\underbrace{N, N, \ldots N}_{b_2~\text{times}}]~|\ldots|~f_{|F|}[\underbrace{N, N, \ldots N}_{b_{|F|}~\text{times}}] ~|~l_1~|~l_2~|\ldots|~l_{|L|}$$

We can imagine the start symbol as representing a (partially constructed) decision tree with a single node that is not yet labeled by a function from $F$ or by a label from $L$. 
Every application of a production rule assigns a function $f_i$ or a label $l_j$ to an unlabeled node $N$. 
If a function $f_i$ is assigned to an unlabeled node, then $b_i$ outgoing edges are created, and $b_i$ new unlabeled nodes are created, one for each new outgoing edge. 
We represent each function $f_i$ in the grammar rules as $f_i[\underbrace{N, N, \ldots N}_{b_i~\text{times}}]$ to indicate that $b_i$ new unlabeled nodes are created, one for each branch in the range of the function $f_i$.  
If a label $l_j$ is assigned to an unlabeled node, no new nodes are created. 
We call decision trees containing unlabeled nodes as partial decision trees, and those not containing unlabeled nodes as complete decision trees. 
Figure~\ref{fig:PDT} gives an example of a partial decision tree.
Unless specified otherwise, by decision trees we will henceforth mean complete decision trees.

We now give the semantics of a decision tree: given an input $x\in \mathcal{I}$, a decision tree $D$ outputs a label $l\in\mathcal{O}$, which we denote by $D(x)$, as per the following rules:
\begin{itemize}
    \item if $D$ is of the form $f_i[D_1, D_2, \ldots D_{b_i}]$ then $D(x):=D_{f_i(x)}(x)$
    \item if $D$ is a leaf labeled $l_j$, then $D(x):=l_j$
\end{itemize}
Note that $D_{f_i(x)}(x)$ is well defined since the range of $f_i$ is $\{1,2,\ldots b_i\}$.

As in~\cite{DBLP:conf/fmcad/TorfahSCAS21},  we assume that we have a node budget $B\in \mathbb{Z}_{>0}$ --- a positive integer --- that bounds the number of internal nodes in our decision trees.
Thus, the class of interpretations $\C{G}$ is restricted to decision trees generated by the grammar $\mathfrak{G}$ with at most $B$ internal nodes. 
We denote the grammar with this restriction by $\mathfrak{G}_{B}$ and the corresponding
class of interpretations by $\C{G}_B$. 
Note, however, that the above is not a real restriction, since the number of semantically different decision trees for a finite set of features/predicates (represented by $f_i \in F$) and labels (represented by $l_j \in L$) is finite. To see this, observe that along any directed path $f_{i_1}\xrightarrow{b_{i_1}}f_{i_2}\xrightarrow{b_{i_1}}\ldots f_{i_j}\xrightarrow{b_{i_j}}\ldots f_{i_{k-1}}\xrightarrow{b_{i_{k-1}}}f_{i_{k}}$ in a decision tree, if there are two nodes with the same function $f_{i_j}=f_{i_k}$, then the second occurrence of $f_{i_k}$ can be removed from the tree, by simply attaching the sub-tree rooted at the $b_{i_j}^{th}$ branch of $f_{i_j}$ to the $b_{i_{k-1}}^{th}$ branch of $f_{i_{k-1}}$. It is easy to see that this doesn't change the semantics since any input $x\in\C{I}$ evaluated along the path already evaluates to $f_{i_j}(x)=f_{i_k}(x)=b_{i_j}$.
Therefore, there are only finitely many semantically different decision trees for a given set of functions and labels, and setting $B$ to the maximum number of internal nodes in this set of semantically different trees allows us to include all semantically different decision trees in the class $\C{G}_B$.

\subsubsection{Explainability and Accuracy Measures}\label{sec:eam}
We now define the explainability measure $\mathcal{E}$ and the correctness measure $\mathcal{C}$ on decision trees.
Our correctness measure $\mathcal{C}:\C{G}_{B}\rightarrow [0,1]$ is a map from the set of decision trees to a number between 0 and 1. 
This gives the accuracy of the decision tree in predicting the output from an input, for samples drawn from the underlying distribution $\Delta(\mathcal{I}\times\mathcal{O})$. 
That is, given a decision tree $D\in \mathcal{G}_B$, its correctness measure $\mathcal{C}(D):=\mathcal{P}_{(i,o)\sim \Delta(\mathcal{I}\times\mathcal{O})}[D(i)=o]$ is the probability of correctly classifying an input-output sample $(i,o)$ from the distribution $\Delta(\mathcal{I} \times \mathcal{O})$. 
We estimate this measure using the Probably Approximately Correct (PAC)~\cite{DBLP:books/daglib/PAC} framework, as done in~\cite{DBLP:conf/fmcad/TorfahSCAS21}.  Specifically, for a given tolerance $\varepsilon$ and confidence $1 -\delta$, where $0 < \varepsilon,\delta < 1$, we first compute the sample complexity, say $\mu(\C{G}, \varepsilon, \delta)$,  of the finite class $\C{G}_B$ of decision trees using standard techniques~\cite{DBLP:books/daglib/PAC}. Then, we draw $\mu(\C{G}, \varepsilon, \delta)$ i.i.d. samples of (input, output) values from the distribution $\Delta(\C{I} \times \C{O})$, and compute the fraction of times the output is correctly predicted by the decision tree $D$ under consideration for these samples. PAC theory then guarantees that with probability at least $1 - \delta$, the computed estimate of the fraction lies within an additive tolerance of $\varepsilon$ of the expected value of the fraction, when i.i.d. samples are drawn from the distribution $\Delta(\C{I} \times \C{O})$.

 
Our explainability measure $\mathcal{E}:\C{G}_B \rightarrow \mathbb{R}_{> 0}$ is a map from the set of decision trees to the set of positive real numbers. Below, we describe a specific explainability measure that assigns higher scores to small decision trees (typically, more explainable than large trees), and also to decision trees that use predicate/features that are more desirable to be included in an interpretation.  Our overall approach is however not restricted to this specific explainability metric, and applies to any metric that can be encoded symbolically (see Section~\ref{sec:SATencoding}).

We assume that each function $f_i$ has an associated (user-provided) weight $w_i\in \mathbb{Z}_+$, that gives us a measure of the desirability of including  $f_i$ in an interpretation. 
Higher weights represent functions that are more desirable to be included. 
Let $W:=\max\limits_{1\leq i\leq |F|}w_i$ be the highest such weight.
Given a decision tree $D$ with $m$ internal nodes containing $i_1$ nodes corresponding to the function $f_1$, $i_2$ nodes corresponding to $f_2$, $\ldots$,  $i_{|F|}$ nodes corresponding to $f_{|F|}$, the explainability measure is given by $\mathcal{E}(D):=(B-m)|W+1|+ \sum\limits_{j=1}^{j=|F|} i_j*w_j$. 
That is, we first prioritize the unused nodes by giving them the highest weights. 
So, lower values of $m$ correspond to more explainable the decision trees. 
This aligns with human intuition, that is, the smaller the decision tree, the easier it is to explain that decision tree.
Then for a given size $m$, we prioritize the functions that are the most desirable to be included in an interpretation. 
This also aligns with human intuition as some functions may be difficult to explain than others.



\section{A Two-Phase Algorithm: Search and Verification}
\label{sec:instantiating_search_verif_proc}

Recall from our earlier discussion that our approach works in two phases. 
In the first phase, we use multi-objective Monte-Carlo Tree Search (MCTS) to solve a multi-objective optimization problem.
Then, in the second phase we use SAT solvers to get local guarantees.
In Section~\ref{sec:MOMCTS} we construct a Multi-Objective Markov Decision Process (MO-MDP). This MO-MDP has (partial or complete) decision trees as states, and uses actions that allow us to grow these decision trees. 
Then, we use Multi-Objective Monte-Carlo Tree Search \cite{DBLP:journals/jmlr/WangS12} to search through the MO-MDP to synthesize best-effort Pareto-optimal decision trees. 
Finally, we give local guarantees on the decision trees synthesized by the MO-MCTS using a Boolean satisfiability solver.

\subsection{Multi-Objective MDP and MCTS}\label{sec:MOMCTS}

In this section, we model the generation of decision trees by the grammar $\mathfrak{G}_B$ as a deterministic MO-MDP, where each state corresponds to a (partial or complete) decision tree, and the production rules of $\mathfrak{G}_B$ determine the actions. 
Specifically, the transitions only have a probability of 0 or 1 depending on the current and next state. 
The probability is 1 for transitions that correspond to the production rule of the chosen action, and 0 otherwise. 
We then assign a multi-objective reward to the states, and apply MO-MCTS~\cite{DBLP:journals/jmlr/WangS12} to synthesize a best-effort approximation to the Pareto-optimal front.

\begin{definition}[Multi-objective MDP]\label{momdp}
    A MO-MDP $M:=$ ($S$, $A$, $T$, $s_0$, $R$) is a tuple, where $S$ is a set of states, $A$ is the set of actions, $T:S\times A \times S\rightarrow [0,1]$ is the transition probability, $s_0$ is the initial state, and $R:S\times A\times S \rightarrow \mathbb{R}^2_{\geq 0}$ is a \textbf{two-dimensional vector-valued reward}.
\end{definition}

We assume that we are given a set $F:=\{f_1, f_2, \ldots f_{|F|}\}$ of functions, a set $L:=\{l_1, l_2, \ldots l_{|L|}\}$ of labels, a budget $B$ bounding the number of internal nodes, and the grammar $\mathfrak{G}_B$ as defined in Section~\ref{sec:problemformulation}. We define the following deterministic MO-MDP $M_{\mathfrak{G}_B}$, where:
\begin{itemize}
    \item Set of states $S:=\{\alpha~|~N\overset{*}{\Rightarrow}_{\mathfrak{G}_B} \alpha \}$
    is the set of all (partial or complete) decision trees generated by the application of zero or more rules from the grammar $\mathfrak{G}_B$ on the initial symbol $N$.

    \item The initial state $s_0$: $s_0=N$ is the initial string $N$, which represents a partial decision tree with just one node corresponding to the non-terminal $N$.
        
    \item Set of actions $A$: $A:=\{(i,j)~|~1\leq i\leq N_{max} \text{ and } 1 \leq j\leq |F|+|L|\}$ is the set of all actions. 
    Here $N_{max}$ is the maximum number of non-terminals of a decision tree in the set $\C{G}_B$.
    Each action is a tuple $(i,j)$ that represents the application of $j^{th}$ production rule of $\mathfrak{G}_B$ on the $i^{th}$ non-terminal node to generate the next (partial or complete) decision tree.
    If the number of non-terminals in the current decision tree is less than $i$, then the transition on this action loops to the same decision tree.

    \item Since we consider a deterministic MO-MDP, the transition on an action $a=(i,j)$ is just the application of $j^{th}$ production rule of $\mathfrak{G}_B$ on the $i^{th}$ non-terminal if it exists, otherwise the transition corresponding to the action just loops back. The transition probabilities are given by the following: \[T((s,a=(i,j),s')):=
            \begin{cases} 
            1, & \begin{aligned} 
                &\text{if the number of non-terminals in $s$ is more}\\
                &\text{than $i$, and applying the $j^{th}$ production rule}\\
                & \text{on $i^{th}$ non-terminal results in $s'$}
            \end{aligned} \\
            1, & \text{if $s=s'$ and there are less than $i$}\\
            & \text{non-terminals in $s$}\\
            0, & \text{otherwise}\\
            \end{cases}\]
    \item A two-dimensional reward is given by the correctness measure in one dimension and the explainability measure in the other.
    For every transition $(s,a,s')$ where $s'$ is not a complete decision tree the reward is defined as $R((s,a,s')):=[0,0]$. Otherwise, if $s'$ is a complete decision tree, the reward is defined as $R((s,a,s')):=[\mathcal{C}(s'), \mathcal{E}(s')]$. i.e. the correctness and explainability measure of the (complete) decision tree $s'$. 
    It is easy to see that we have a sparse setting for the reward.
\end{itemize}

\begin{figure}[t]
    \centering
    \includegraphics[width=0.5\linewidth]{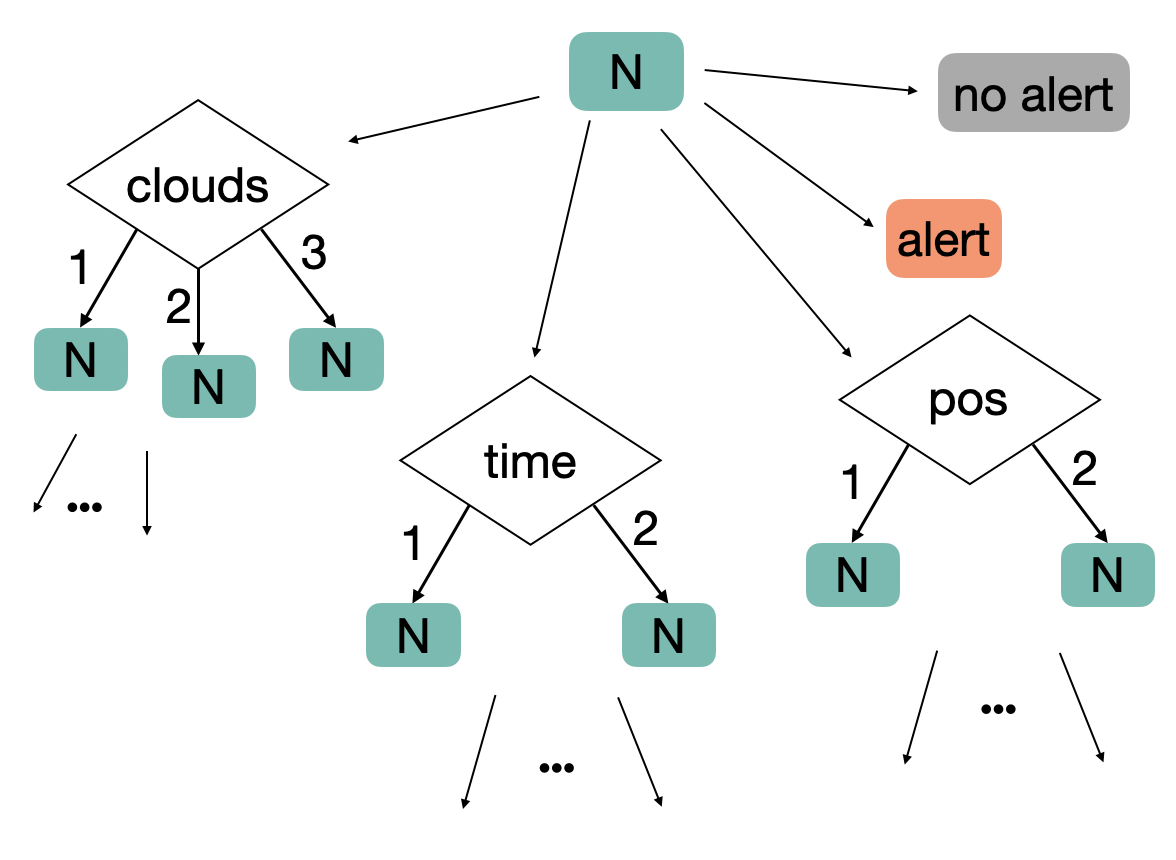}
    \caption{Deterministic transitions in the MO-MDP }
    \label{fig:momdp}
\end{figure}

Figure~\ref{fig:momdp} shows some deterministic transitions from the initial node in an MO-MDP modeling one of our benchmarks (TaxiNet), where \emph{clouds, time, pos} are elements of $F$, and \emph{alert, noalert} are labels in $L$.
The MO-MCTS algorithm developed in \cite{DBLP:journals/jmlr/WangS12} works for deterministic MO-MDPs, and is therefore  suitable use in our approach.
The MO-MCTS procedure of \cite{DBLP:journals/jmlr/WangS12} generates a search tree and maintains a set of best-effort approximations of Pareto-optimal interpretations over the search tree.
This procedure has three phases. 
In the first phase, the  procedure iteratively chooses an action until it arrives at a (search tree) node that has all unexplored actions, or until a Progressive Widening heuristic is triggered on a node with unexplored actions. This heuristic initially restricts the number of child nodes allowed for exploration. For each node, it only allows a limited number of child nodes for exploration at the start, and gradually increases this number as the visits to the parent node increase. This helps in managing the exploration initially by focusing on a smaller subset of actions. 
During the first phase, the actions chosen are those that maximize a quantity dependent on the hypervolume indicator~\cite{DBLP:conf/ppsn/ZitzlerT98} 
of the upper confidence bound, subtracted with an $L_2$-norm of the perspective projection of the upper confidence bound.
Then it chooses an unexplored action using another heuristic that builds on the Rapid Action Value Estimation (RAVE) heuristic~\cite{DBLP:conf/icml/GellyS07} 
 by maximizing over all unexplored actions the $L_2$-norm of the difference between the perspective projection of the RAVE heuristic with itself.
The second phase starts once an unexplored action is chosen. 
In this phase,  the MO-MCTS procedure randomly selects actions until a terminal state is reached.
Then in the third and final phase, the multi-dimensional rewards are collected and the quantities required for the heuristics are updated by backtracking.

The MO-MCTS procedure maintains a set of best-effort approximations of Pareto-optimal interpretations that it keeps on improving in each iteration. Lemma~\ref{lem:l1} below  quantifies this formally.
\begin{lemma}\label{lem:l1}
    Let $PO_i$ be an incomparable set of best-effort Pareto optimal points in the MO-MCTS procedure at iteration $i$. Then for every iteration $j>i$ and for every point $p\in PO_i$ there is a point $p'\in PO_j$ such that $p\preceq p'$.  
\end{lemma}

\subsection{SAT Encoding}
\label{sec:SATencoding}
In this section, we describe the verification procedure for the interpretations obtained from the MO-MCTS procedure. 
We assume that the MO-MCTS procedure outputs a set $S:=\{(D,\C{C}(D),\C{E}(E))\}$ of decision trees along with their correctness and explainability measures.
Let $\delta_c$ and $\delta_e$ be the user-provided correctness and explainability slacks respectively.
For each interpretation $(D, \C{C}(D), \C{D}(D))$ in $S$, this phase searches for an interpretation $D' \in \mathcal{G}_B$ s.t. $(\C{C}(D)+\delta_c, \C{E}(D)+\delta_e)\succeq (\C{C}(D'), \C{E}(D'))\succ (\C{C}(D), \C{E}(D))$ by encoding the problem as one of Boolean satisfiability, and by using a Boolean satisfiability solver.  If it finds such a $D'$, then it replaces $D$ with $D'$ in the set $S$, and repeats the above search using the satisfiability solver.  Otherwise, it declares $D$ as locally PO w.r.t $\delta_c$ and $\delta_e$ and moves it from $S$ to a set $S'$. This phase also filters out the decision trees in $S'$ that are already Pareto-dominated by other locally PO trees in $S'$. Thus, if allowed to run till completion, this phase computes a set $S'$ of locally Pareto-optimal (w.r.t. $\delta_c$ and $\delta_e$) decision trees that are Pareto-incomparable, such that each tree in  $S$ is either present in $S'$ or is Pareto-dominated by a tree in $S'$.

To search for an interpretation $D'$ satisfying $(\C{C}(D)+\delta_c,~\C{E}(D)+\delta_e)$ $\succeq$ $(\C{C}(D'),~\C{E}(D'))\succ (\C{C}(D),~\C{E}(D))$, we use the encoding ideas in~\cite{DBLP:conf/fmcad/TorfahSCAS21}.  Specifically, we construct a Boolean formula $\Phi(X, Y, Z, W)$, where $X, Y, Z, W$ are binary encodings of integers, that is satisfiable if and only if there is a decision tree $D' \in \C{G}_B$ such that $(X,Y)\prec (\C{C}(D'),\C{E}(D'))\preceq (Z, W)$.  Moreover, the formula is constructed such that the interpretation $D'$ can be obtained directly from the satisfying assignment of $\Phi$. 
Once $\Phi$ is obtained, we invoke a Boolean satisfiability solver on $\Phi(\C{C}(D),\C{E}(D),\C{C}(D)+\delta_c, \C{E}(D)+\delta_e)$ to find the desired $D'$. If the formula is unsatisfiable, then we know that $D$ is LPO w.r.t $\delta_c$ and $\delta_e$. 

Motivated by the encoding ideas in~\cite{DBLP:conf/fmcad/TorfahSCAS21}, we obtain the formula $\Phi(X, Y, Z, W)$ is as a conjunction of four sub-formulas
$$\Phi(X, Y, Z, W):=\Phi_{syntax}\land \Phi_{corr}(X, Z) \land \Phi_{exp}(Y, W) \land \Phi(X, Y).$$
Here, $\Phi_{syntax}$ encodes a syntactic restriction that allows only those decision trees generated by the grammar $\mathfrak{G}_B$. Similarly,  $\Phi_{corr}(X, Z)$ encodes a correctness restriction that is satisfiable if there is a decision tree $D'$ such that $X \leq \C{C}(D')\leq Z$. The formula $\Phi_{exp}(Y, W)$ is s.t. it is satisfiable iff there is a decision tree $D$ such that $Y \leq \C{E}(D) \leq W$. Finally, the fourth component $\Phi(X,Y)$ is satisfiable iff there is a decision tree $D$ such that the correctness and explainability measures dominate $(X,Y)$, i.e., $(X,Y)\prec (\C{C}(D'), \C{E}(D'))$. 
We explain the encoding in more detail in Appendix~\ref{A1}. 
The correctness of the encoding is formalized below.
\begin{lemma}\label{lem:bool}
    There is a Boolean formula $\Phi(c,e,c+\delta_c, e+\delta_e)$ that is satisfiable if and only if there is a decision tree $D' \in \mathcal{G}_B$ such that $(c,e)\prec (\C{C}(D'),\C{E}(D'))\preceq (c+\delta_c,e+\delta_e)$.
\end{lemma}

\subsection{Overall Procedure}

We now integrate the ideas in the previous two sections to come up with an anytime two-phase algorithm. 
%
Algorithm~\ref{alg:LPO2} takes as input the timeout $T_{\text{MO-MCTS}}$ of the first phase involving MO-MCTS, the overall timeout $T_{\text{overall}}$, and the slacks $\delta_c$ and $\delta_e$.  The algorithm runs in two phases.
In the first phase, it executes the MO-MCTS procedure with a timeout $T_{\text{MO-MCTS}}$. 
Since MO-MCTS always maintains a set of best-effort Pareto-optimal points internally, we collect all these best-effort solutions in a set $S$ once the time budget $T_{\text{MO-MCTS}}$ is exhausted.  Specifically, we assume that at the end of the first phase, each element in set $S$ is a decision tree $D \in \mathcal{G}_B$ along with the tuple $(\C{C}(D), \C{E}(D))$.

\begin{algorithm}[t]
\caption{Anytime LPO Interpretation Synthesis}\label{alg:LPO2}
\KwIn{
  Timeouts $T_{\text{MO-MCTS}}$ and $T_{\text{overall}}$ of MO-MCTS and overall algorithm\\
  ~~~~~~~~~~~~Slack $\delta c$ of correctness measure $\C{C}$\\
  ~~~~~~~~~~~~Slack $\delta e$ of explainability measure $\C{E}$
}
\KwOut{Set $S'$ of incomparable interpretations that are LPO w.r.t. $\delta c$ and $\delta e$\\
~~~~~~~~~~~
Set $S$ of best-effort interpretations}
\DontPrintSemicolon
\;
\tcc{Execute the first phase of MO-MCTS}
$S \gets \texttt{execute\_with\_timeout}(\text{MO-MCTS}, T_{\text{MO-MCTS}})$\;
\;
\tcc{Execute the second phase of verification}
$S'\gets \{\}$\;
\;
\While{$S\neq \emptyset$ ~or~ $T_{\text{overall}}$ \text{not exceeded}}{
\;
$(D, (c,e)) \gets S.pop()$ \tcp{fetch and remove an element $D,(c,e)$ from $S$}\;
(status, assignment) = \texttt{Check\_SAT}$(\Phi(c,e,c+\delta_c, e+\delta_e), T_{\text{overall}})$ \;
    \If{ \textrm{status is unsatisfiable}}{
        $S' \gets S'\cup \{(D,(c,e))\}$\;
        Remove Pareto-dominated interpretations from $S'$\;
    }
    \ElseIf{\textrm{status is satisfiable}}{
        \tcc{the satisfying assignment encodes a decision tree $D'$ with measures $(c',e')$ s.t. $(c,e)$ $\prec$ $(c',e')$ $\preceq$ $(c+\delta_c, e+\delta_e)$}
        $(D', (c',e'))$ $\gets$ extract\_decision\_tree(assignment)\;
        $S$ $\gets$ $S$ $\cup$ $\{(D',(c',e'))\}$\;
        Remove Pareto-dominated interpretations from $S$\;
    }
    \Else{
    \tcc{$T_{\text{overall}}$ exceeded during \texttt{Check\_SAT} call}
    $S$ $\gets$ $S \cup \{ (D, (c, e)) \}$\;
    }
}

\Return{$S'$ and $S$}\;
\end{algorithm}

In the second phase, Algorithm~\ref{alg:LPO2} maintains a set $S'$ of confirmed locally Pareto-optimal interpretations, and initializes $S'$ to the empty set.  It then
iterates through each tree $D$ in $S$, and invokes the \texttt{Check\_SAT} function on the Boolean formula $\Phi(\C{C}(D),\C{E}(D),\C{C}(D)+\delta_c, \C{E}(D)+\delta_e)$. 
Function \texttt{Check\_SAT} checks if the formula fed to it as input is satisfiable. 
If so, it returns a decision tree $D'$, along with $(\C{C}(D'), \C{E}(D'))$ such that 
$(\C{C}(D), \C{E}(D)) \prec (\C{C}(D'), \C{E}(D')) \preceq (\C{C}(D) + \delta_c, \C{E}(D) + \delta_e)$.  Since $D'$ Pareto-dominates $D$, we replace the entry $(D, (\C{C}(D), \C{E}(D)))$ in $S$ by $(D', (\C{C'}(D), \C{E}(D')))$.
If, on the other hand, \texttt{Check\_SAT} reports that $\Phi(\C{C}(D),\C{E}(D),\C{C}(D)+\delta_c, \C{E}(D)+\delta_e)$ is unsatisfiable, we know that $D$ is LPO w.r.t. $\delta_c$ and $\delta_e$, and add $(C, (\C{C}(D), \C{E}(D)))$ to the set $S'$, while ensuring that no two interpretations in $S'$ Pareto-dominate each other.  If the overall timeout $T_{\text{overall}}$ is exceeded while \texttt{Check\_SAT} is in execution, we assume that \texttt{Check\_SAT} is forcibly terminated and the status returned indicates timeout.  In such cases, we retain $(D, (\C{C}(D), \C{E}(D)))$ in the set $S$.

The second phase of Algorithm~\ref{alg:LPO2} iterates until the overall timeout $T_{\text{overall}}$ is reached or the set $S$ becomes empty.  In either case, the algorithm returns the sets $S'$ and $S$, with the guarantee that $S'$ contains incomparable LPO interpretations w.r.t. $\delta_c$ and $\delta_e$.  The interpretations in $S$, however, are simply best-effort interpretations that may not be LPO w.r.t. $\delta_c$ and $\delta_e$.

\begin{theorem}\label{thm:algo2}
    Suppose  Algorithm~\ref{alg:LPO2} outputs sets $S'$ and $S$ of decision trees along with their associated measures on termination. Then, every decision tree $D$ in $S'$ is LPO w.r.t. $\delta_c$ and $\delta_e$. Furthermore, for every decision tree $D$ in $S$, there exists at least one tree $D'$ output by MO-MCTS s.t. $(\C{C}(D'), \C{E}(D')) \preceq (\C{C}(D), \C{E}(D))$.
\end{theorem}
Note that Theorem~\ref{thm:algo2} holds regardless of whether Algorithm~\ref{alg:LPO2} terminates due to $S$ becoming empty or $T_{\text{overall}}$ being exceeded. It is also easy to verify that Algorithm~\ref{alg:LPO2} is guaranteed to output at least one decision tree (in either $S$ or $S'$) if the first phase of MO-MCTS does not return an empty set.  This suggests the following corollary.
\begin{corollary}
With sufficiently large overall timeout $T_{\text{overall}}$ and with sufficiently large values of $\delta_c$ and $\delta_e$, Algorithm~\ref{alg:LPO2} is guaranteed to find at least one PO interpretation.
\end{corollary} 

\section{Experiments}
\label{sec:exp}
\newcommand{\tool}{\textsc{ALPO}}

We have implemented the MO-MCTS procedure of~\cite{DBLP:journals/jmlr/WangS12} adapted to our setting, and the verification algorithm described above in a prototype tool called {\tool} available at \href{}{https://github.com/anirjoshi/ALPO}. 
All our experiments were run on an Apple Mac M2 Pro with 32 GB memory.
In the experiments, we set the timeout $T_{\text{MO-MCTS}}$ for MO-MCTS to be exactly half of the overall timeout $T_{\text{overall}}$.  We experimented with two different values of the overall timeout $T_{\text{overall}}$, viz. 5 mins and 20 mins.
We used the Kissat SAT solver~\cite{BiereFallerFazekasFleuryFroleyksPollitt-SAT-Competition-2024-solvers} to solve the Boolean satisfiability queries in the verification phase of Algorithm~\ref{alg:LPO2}.
In the first phase involving MO-MCTS, we restricted the actions of MO-MDP to only allow those actions that involve the first non-terminal.
Without this restriction, we empirically observed that the action set becomes very large. 
We also note that the restriction does not reduce the expressivity, as the actions may be applied in any order. For example, in Figure~\ref{fig:PDT}, actions for any of the two non-terminals can be taken first. This restriction just enforces an ordering on the actions by always selecting an action corresponding to the first non-terminal. 

For each state, we also skipped explore the self-looping actions, since they do not contribute towards exploration. 
Finally, from the initial state of the MO-MDP we do not allow actions involving labels in the set $L$, since allowing such actions would give rise to single-node decision trees that classify every input to the same label.
%
We compare our work with the tool Synplicate developed as part of ~\cite{DBLP:conf/fmcad/TorfahSCAS21}, which is guaranteed to synthesize the full Pareto-optimal curve of interpretations, when it does terminate (which as we show is not always the case). Since Synplicate considers interpretations as decision diagrams, we modified it slightly to restrict the interpretations to decision trees.  We used the same timeout $T_{\text{overall}}$ for Synplicate as used by {\tool}.  


\subsubsection*{Benchmarks}

For our benchmarks, we considered a set of 10 benchmarks from the UCI repository~\cite{uci}, 3 custom-made benchmarks, and 5 randomly generated benchmarks. In this section, we present the four most relevant benchmarks namely AutoTaxi, Balance Scale~\cite{balance_scale_12}, Car Evaluation~\cite{car_evaluation} and Yeast~\cite{yeast_110}. The remaining are detailed in~Appendix~\ref{A2}.

All of our benchmarks have categorical outputs. The AutoTaxi benchmark, adapted from \cite{DBLP:conf/cav/FremontCMOS20,DBLP:conf/rv/TorfahJFS21}, is a decision module that predicts whether a perception module of an airplane behaves correctly under certain environment conditions. It uses the following features: time of day, types of clouds, and initial position of the airplane on the runway.
The implementation of this module is a decision tree based on data collected from 200 simulations, using the XPlane (\url{https://www.x-plane.com}) simulator.

The other benchmarks Balance Scale, Car Evaluation, and Yeast are all collected from the UCI Machine Learning Repository~\cite{uci}.
These benchmarks are available as datasets. 
We first train these datasets on a fully connected neural network with four hidden layers containing seven neurons in each layer, and a ReLU activation function for each neuron.
This neural network then serves as our black-box model.
For the synthesis procedure, we restrict the space of decision trees to only those with size less than six internal nodes. 
The names of functions for the decision tree are the same as the names of features in the dataset.
For every categorical feature, we assign a number to each category, and the feature function in the decision tree just outputs the number according to the category.
For every numerical feature $\mathbf{f}$, we have a function $f$ that has three output buckets. 
This function simply divides the range of the feature into three equal regions. 
Let $\mathbf{f}_{max}$ and $\mathbf{f}_{min}$ be the maximum and minimum values of $\mathbf{f}$ in the dataset. Then $f\colon\mathbb{R}\rightarrow\{0,1,2\}$ is defined as:
\[f(i):=
    \begin{cases} 
            0, & \text{if } i <  \mathbf{f}_{min} + \frac{\mathbf{f}_{max}-\mathbf{f}_{min}}{3} \\
            1, & \text{else if } i < \mathbf{f}_{min} + 2*\frac{\mathbf{f}_{max}-\mathbf{f}_{min}}{3}\\
            2, & \text{otherwise}\\
    \end{cases}\]

In Table~\ref{tab:my_label}, we list the features used in these four benchmarks.
We plot the results of executing {\tool} and Synplicate on these benchmarks in Figure~\ref{fig:all_datasets}. We first discuss the results with an overall timeout of 20 minutes.
(10 mins for MO-MCTS, and 10 mins for the second phase in {\tool}).
The $(\delta_c,\delta_e)$ windows for AutoTaxi, Car Evaluation, Balance Scale, and Yeast are $(0.023,5)$, $(0.018,5)$, $(0.021,5)$ and $(0.017,5)$,  respectively. The $\delta_c$ values are obtained as $\frac{10}{K}$, where $K$ is the total number of samples used for a benchmark. 
We use $\epsilon=0.25$ and $\delta=0.1$ for the PAC guarantees, without making any realizability assumption.

\begin{table}[t]
    \centering
    \begin{tabular}{|c|c|c|c|}
      \hline
      \textbf{Benchmark Name} & \textbf{Features} & \textbf{Weights} &  \textbf{Output Branches}\\
      \hline
       AutoTaxi  & clouds, day time, init pos & 1, 4, 3 & 6, 3, 4\\
       Balance Scale & left distance, left weight & 3, 3 & 3, 3\\
                     & right distance, right weight & 3, 3 & 3, 3\\
       Car Evaluation & buying, doors, lug boot & 3, 3, 3 & 4, 4, 3\\
                      & maint, persons, safety & 3, 3, 3 & 4, 3, 3\\
       Yeast & alm, erl, gvh, mcg, mit & 3, 3, 3, 3 & 3, 3, 3, 3\\
             & nuc, pox, vac & 3, 3, 3 & 3, 3, 3\\
       \hline
    \end{tabular}
    \caption{List of feature names, weights and output branches in all benchmarks}
    \label{tab:my_label}
\end{table}

\subsubsection{The research questions}
We address the following three research questions.
\begin{itemize}
    \item RQ1: Are locally Pareto-optimal solutions given by our approach close to the actual (global) Pareto-optimal solutions?
    \item RQ2: Does the MO-MCTS procedure output a good approximation of the Pareto-optimal curve?
    \item RQ3: Does our procedure scale empirically in comparison with Synplicate?
\end{itemize}

\subsection{Our Results}
\noindent{ \bf RQ1: From local Pareto-optimality to global Pareto-optimality.} Our experiments reveal that the search for locally Pareto-optimal solutions around the solutions of the MO-MCTS procedure often gives us globally Pareto-optimal solutions. 
We use the AutoTaxi and the Balance Scale benchmarks to demonstrate this.  Indeed, these are the only benchmarks (among the four being discussed here) where the Synplicate tool terminates within 20 mins, and hence gives the Pareto-optimal curve, allowing us to perform this comparison. 



\begin{figure}[tbh!]
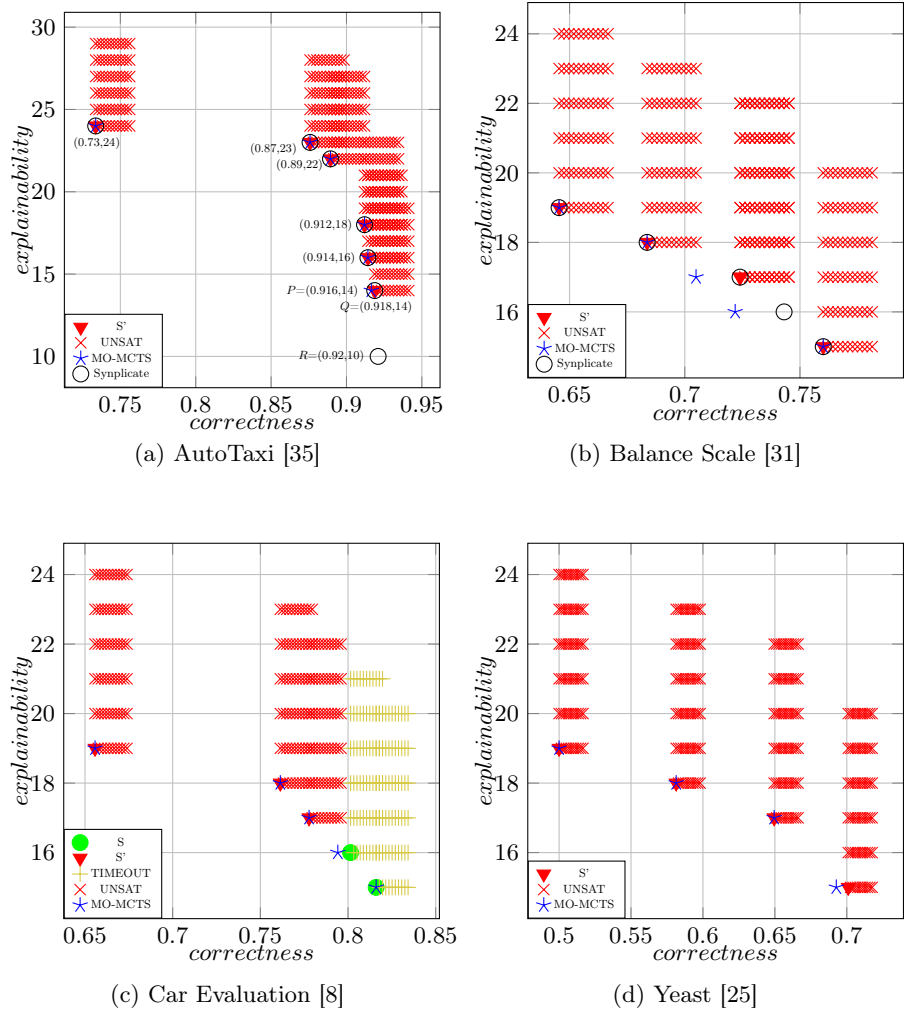

    \centering
            \centering
            \begin{subfigure}[b]{0.49\textwidth}
                \centering
                    \include{figures/autotaxi}
                \vspace{-2.9em}
                \caption{ AutoTaxi~\cite{DBLP:conf/fmcad/TorfahSCAS21}}
                \label{fig:autotaxi}
            \end{subfigure}
            \begin{subfigure}[b]{0.49\textwidth}
                \centering
                \include{figures/balance_scale}
                \vspace{-2.5em}
                \caption{Balance Scale~\cite{balance_scale_12}}
                \label{fig:balance_scale}
            \end{subfigure}

            \vspace{0.9cm}

            \begin{subfigure}[b]{0.49\textwidth}
                \centering
                \include{figures/car_evaluation}
                \vspace{-2.5em}
                \caption{Car Evaluation~\cite{car_evaluation}}
                \label{fig:car_eval}
            \end{subfigure}
            \begin{subfigure}[b]{0.49\textwidth}
                \centering
                \include{figures/yeast}
                \vspace{-2.5em}
                \caption{Yeast~\cite{yeast_110}}
                \label{fig:yeast}
            \end{subfigure}
        \caption{Visualization of  results: For each benchmark, correctness measure is plotted on X-axis and explainability measure on Y-axis.  Blue stars indicate MO-MCTS results. Red inverted triangle indicate LPO interpretations found by {\tool}. Red crosses indicate absence of interpretations with corresponding explainability-accuracy measures. Green solid circles indicate points that dominate an interpretation obtained from MO-MCTS, but not verified to be LPO. Yellow pluses indicate that {\tool} timed out when checking for LPO here. Synplicate results are shown with hollow circles.}
    \label{fig:all_datasets}
    \vspace{-0.2cm}
\end{figure}

\begin{minipage}[t]{0.45\textwidth}
\centering
\includegraphics[width=\linewidth]
{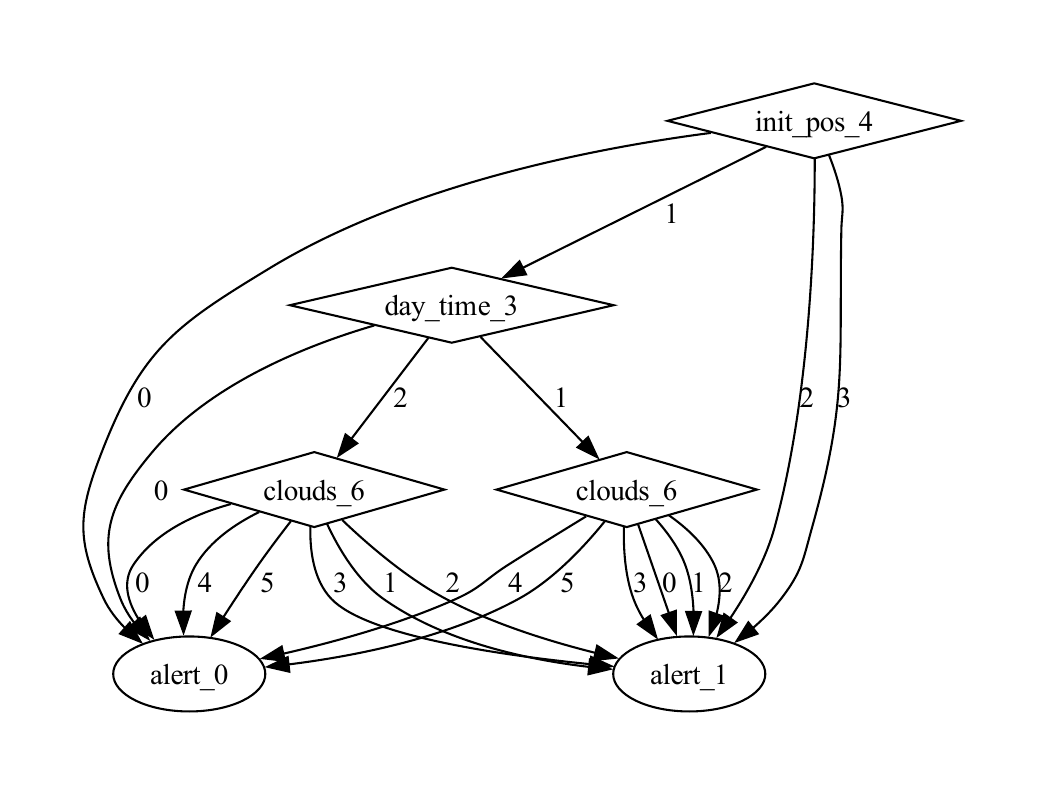}
  \captionof{figure}{Decision tree corresponding to point $P$ in Figure~\ref{fig:autotaxi} from the first phase of MO-MCTS with correctness and explainability measures $0.916$ and $14$ respectively.}
  \label{fig:mo_mcts_po}
\end{minipage}%
\hfill
\begin{minipage}[t]{0.45\textwidth}
  \hspace{-5em}
\includegraphics[width=1.3\linewidth]{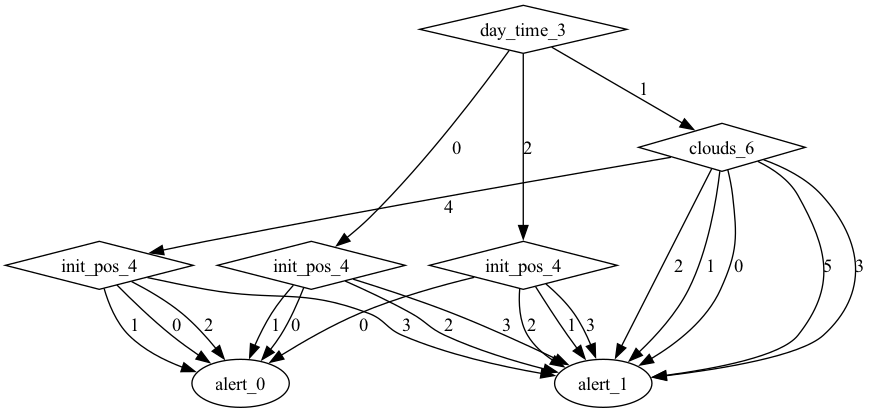}
  \captionof{figure}{Decision tree corresponding to point $Q$ from the second phase of verification procedure with correctness and explainability $0.918$ and $14$ resp.}
  \label{fig:syn}
\end{minipage}

\medskip

\smallskip


In the AutoTaxi benchmark, we observe from the Pareto-optimal curve in  Figure~\ref{fig:autotaxi} that all decision trees except one [$P$=(0.916,14)] from the MO-MCTS procedure are Pareto-optimal. 
Also, we are able to discover a Pareto-optimal decision tree $Q$ close to $P$ = (0.916, 14) in the second phase. 
We show the decision trees representing points $P$ and $Q$ in Figures~\ref{fig:mo_mcts_po} and \ref{fig:syn} respectively. 
\smallskip 

\noindent{\bf RQ2: MO-MCTS is a good approximation of the Pareto-optimal curve}  
The plots in Figure~\ref{fig:autotaxi} and Figure~\ref{fig:balance_scale} corresponding to the AutoTaxi and Balance Scale benchmarks, respectively, provide evidence that the MO-MCTS procedure gives a good approximation of the (actual) Pareto-optimal curve, as given by Synplicate. 
The MO-MCTS procedure outputs eight of the combined twelve Pareto-optimal points.  
We cannot use the other two benchmarks to answer this question, because the Synplicate tool times out before outputting even one Pareto-optimal point.\smallskip 

\begin{figure}
\centering
\includegraphics[width=0.75\linewidth]{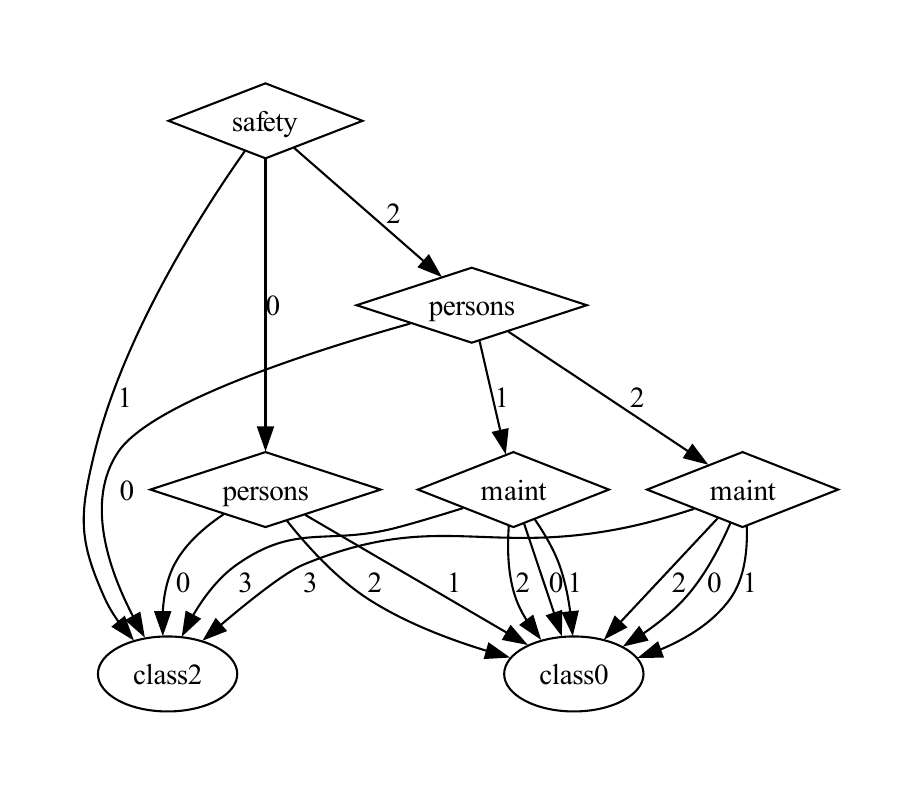}
  \captionof{figure}{Decision tree corresponding to the most accurate best-effort Pareto-point from the MO-MCTS for Car Evaluation benchmark.}
  \label{fig:mom_mcts_car_eval}
\end{figure}

\noindent{\bf RQ3: Performance of {\tool} vs Synplicate} Finally, we address whether our procedure performs better than Synplicate on a wider set of benchmarks. We answer this affirmatively by looking into the two benchmarks: Car Evaluation and Yeast. 
More evidence is presented in Appendix~\ref{A2} via other benchmarks. 
As can be observed from the results of the Car Evaluation benchmark in Figure~\ref{fig:car_eval} and the Yeast benchmark in Figure~\ref{fig:yeast}, the Synplicate tool cannot output even a single Pareto-optimal point for these benchmarks. Since the number of feature nodes in these benchmarks is more than those of the other two, this increases the size of the search space significantly.  We believe this eventually causes Synplicate to choke. However, {\tool} outputs at least four decision trees for each benchmark.
  Figure~\ref{fig:mom_mcts_car_eval} shows the decision tree correponding to highest accuracy decision tree synthesized by this approach. 
Moreover, for each of the two benchmarks, we are able to verify that two decision trees are locally Pareto-optimal. Therefore, using our procedure, we are able to synthesize locally Pareto-optimal decision trees even in the cases where synthesizing a Pareto-optimal set is practically computationally intensive.


\clearpage

\newpage

In summary, our results answer RQ1-3 comprehensively and show the effectiveness of {\tool}.   We also 
conducted experiments on several other benchmarks (see Appendix~\ref{A2}), and repeated the experiments on the four benchmarks 
presented above with a 5 mins timeout (see Appendix~\ref{A3}). 

  Our results consistently support the conclusions w.r.t. RQ1-3 made above.  In particular, when running with timeout of 5 mins,
 {\tool} gives fewer (but non-empty) LPO interpretations than with 20 mins timeout, and hence degrades gracefully. This shows the value of {\tool} as an anytime LPO interpretation synthesis algorithm.

\section{Conclusion}
In this paper, we have considered an integrated multi-objective MCTS-based search algorithm with SAT-based verification to synthesize Pareto-optimal interpretations with locally optimality guarantees. Our experimental results show that the interpretations we generate are close to globally Pareto-optimal illustrating the practical significance of the approach. As future work, 
it would be interesting to use the solutions from the SAT solver to improve search, leading to even better integration between these approaches. Future advancements in methods for MCTS-based search will directly transfer to our approach.

\bibliographystyle{splncs04}
\bibliography{references}

\appendix


\section{Verification using Boolean Satisfiability}
\label{A1}
We describe below the construction a boolean formula $\Phi(X, Y, Z, W)$ that is satisfiable if and only if there is a decision tree $D \in \mathcal{G}_B$ such that $(X,Y)\prec (\C{C}(D), \C{E}(D))\preceq (Z,W)$.  Specifically, we are interested in instances of the formula of the form $\Phi(c,e,c+\delta_c, e+\delta_e)$, where $(c, e)$ gives the accuracy and explainability measure of a known decision tree, and $\delta_c, \delta_e$ are user-provided slacks for accuracy and explainability respectively.  We build on ideas for a similar encoding, used in~\cite{DBLP:conf/fmcad/TorfahSCAS21}. 

We assume a set of functions $F:=\{f_1, f_2, \ldots, f_{|F|}~|~f_i:\mathcal{I}\rightarrow \{1,2,\ldots b_i\}\}$ for the decision tree and the set $L:=\{l_1, l_2, \ldots l_{|L|}\}$ of labels. Let $b_{max}:=\max\limits_{1\leq i\leq |F|} b_i$. We upper-bound the internal nodes by $B$. We will assume the correctness measure defined in Section~\ref{sec:eam}, and assume $\mu(\C{G},\varepsilon, \delta)$ samples $M:=\{(i,o)~|\Delta(\C{I}\times \C{O})\}$ from the input-output distribution. For the explainability measure we assume a user-provided weight $w_i>0$ with every function $f_i$, which determines the explainability of the function. We also assume that there exists a topological ordering over the nodes of every decision tree. It is fine if we do not know this ordering for every decision tree. Only the existence of an ordering will be used in the encoding.

The encoding for $\Phi(c,e,c+\delta_c, e+\delta_e)$ is a conjunction of four components
$$\Phi(c,e,c+\delta_c, e+\delta_e):=\Phi_{syntax}\land \Phi_{corr}(c,c+\delta_c) \land \Phi_{exp}(e,e+\delta_e) \land \Phi(c,e)$$

\subsubsection{$\Phi_{syntax}$} As explained earlier, we modify the encoding from \cite{DBLP:conf/fmcad/TorfahSCAS21} to syntactically restrict the space to decision trees. We now describe the combined encoding:
 
 We let the variable $\lambda_{i,p}$ encode that the $i^{th}$ node in the decision tree is assigned a function $p$. Hence the number of such $\lambda$ variables is $B\times |F|$. We first encode that each node is assigned to exactly one function. 
 Let $E_i:= \{\lambda_{i,p}~|~p\in F\}$, then
 $$F_1:=\bigwedge \limits_{1\leq i\leq B} Exactly\_One(E_i)$$ where $Exactly\_One$ just encodes a boolean constraint that in every satisfying assignment exactly one of the variables in the set is true and all others are false. $F_1$ just encodes that every node is assigned to exactly one function.

 We now let variable $\tau_{i,c,j}$ encode that the $i^{th}$ node has an edge to the $j^{th}$ node or the label $j$, labelled $c$. Since there is a topological ordering, we assume that $i<j$. There are $\frac{B\times (B-1)}{2}\times (b_{max}+|L|)$ such variables.
 We then create a sub-formula which encodes that from every node there is exactly one outgoing edge corresponding to every label. We let $E_{i,c}:=\{\tau_{i,c,j}~|~i<j\leq B\}$ $$G_{i,c}:=Exactly\_One(E_{i,c})$$
 Then we encode the constraint that if a node is assigned a function, then exactly every outgoing edge has a unique label in the range of that function.
 $$F_{i,p}:=\lambda_{i,p}\rightarrow \big(\bigwedge \limits_{1\leq c \leq b_p} G_{i,c} \big)\land \big(\bigwedge\limits_{b_p<c\leq b_max}\lnot \tau_{i,c,j}\big)$$

 Finally we conjoin this over all nodes and all functions:
 $$F_2:=\bigwedge\limits_{\substack{1\leq i\leq B \\ p\in F}} F_{i,p} $$

 Now, we restrict that every node must have at most one incoming edge. For this let $H_{j}:=\{\tau_{i,c,j}~|~1\leq i < j,~1\leq c\leq b_{max}\}$
 $$F_3:=\bigwedge\limits_{ 1 < j\leq B} At\_Most\_1(H_{j})$$

$$\Phi_{syntax}:=F_1 \land F_2 \land F_3$$

\subsubsection{$\Phi_{corr}(c+\delta_c)$} We now encode the leaves $m_{l,i}$ encodes that the $i^{th}$ sample has output label $l$. We assign them True or False depending on if the label on the $i^{th}$ sample is $l$

\[m_{l,i}:=
    \begin{cases} 
            T, & \text{if } o_i=l \\
            F, & \text{otherwise}\\
\end{cases}\]

Then we let the variable $m_{i,j}$ encode the constraint that the $i^{th}$ node in the decision tree outputs correct label for the $k^{th}$ sample correctly, and encode this constraint with the formula:
$$M_{i,k}:=m_{i,k}\leftrightarrow \big( 
 			\bigvee \limits_{p\in F} \bigvee \limits_{c\leq b_p} ~
 			\lambda_{i,p}  \wedge  func(M_k, p, c)  \wedge \\
 			\bigwedge \limits_{\substack{j \in \{i+1,\dots,B\} \cup\\  L}}
            ( \tau_{i,c,j} \rightarrow m_{j,k)})
 			~\big)$$
            where $func(M_k, p, c)$ is $T$ iff the $k^{th}$ sample outputs $c$ on function $p$, i.e, $p(M_k)=c$.
We do this for all samples and have the following $$F_4:=\bigwedge \limits_{\substack{1\leq i\leq B \\ 1\leq k\leq |M|}} M_{i,k}$$

We use the symbol $C$ to denote $C:=\sum_{k} m_{1,k}$

For bounding the correctness measure we use cardinality constraints on $m_{1,k}$ variables:
$$\Phi_{corr}(c,c+\delta_c):=Encode(C\leq c+d_c)\land Encode(c\leq C)\land F_4$$

\subsubsection{$\Phi_{exp}(e+\delta_e)$} We now encode the explainability constraints. Where we have a variable $u_i$ which encodes that the $i^{th}$ node is reachable from the initial node, and have the following constraint:

$$F_5:=u_1 \wedge \bigwedge \limits_{2\leq i \leq B} u_i \leftrightarrow (\bigvee \limits_{\substack{1\leq c \leq c_{\max}, \\1\leq i'<i}} \tau_{i',c,i} \wedge u_{i'}) $$

Then we have another variable $\overline{u_i}$ which encodes that the $i^{th}$ node is not used

$$F_6:=  \bigwedge \limits_{1\leq i \leq B} \lnot u_i \leftrightarrow \overline{u_i} $$

Finally we also have a variable $\lambda'_{i,p}$ which encodes that the $i^{th}$ node is used and has the function $p$ assigned to it which we encode using the following formula:
$$F_7:=\bigwedge \limits_{\substack{1\leq i\leq B\\ p\in F}} \lambda'_{i,p}\leftrightarrow (\lambda_{i,p}\land u_i)$$

Let us use $E$ to denote the term $E:=\sum\limits_{\substack{1\leq i\leq B\\ p\in F}}w_p\lambda'_{i,p} + \sum\limits_{1\leq i\leq B}(W_{max}+1)\overline{u_i}$

Now we encode the formula $$\Phi(e,e+\delta_e):=F_5 \land F_6 \land F_7 \land Encode (E\leq e+\delta_e) \land Encode(e\leq E)$$

\subsubsection{$\Phi(c,e)$} $$\Phi(c,e):=\big(Encode(e\leq E)\land Encode(c<C)\big) \lor \big(Encode(e<E)\land Encode(c\leq C) \big)$$

\section{Extended Experiments}\label{A2}

We present below comprehensive experimental results on all the benchmarks considered in our work.  Only four of these were presented in the main paper due to lack of space.  For all the experiments below, we used a total timeout of $20$ mins (10 mins for MO-MCTS and 10 mins for the second phase of {\tool}).

\begin{figure}
  \includegraphics[width=1.2\linewidth]{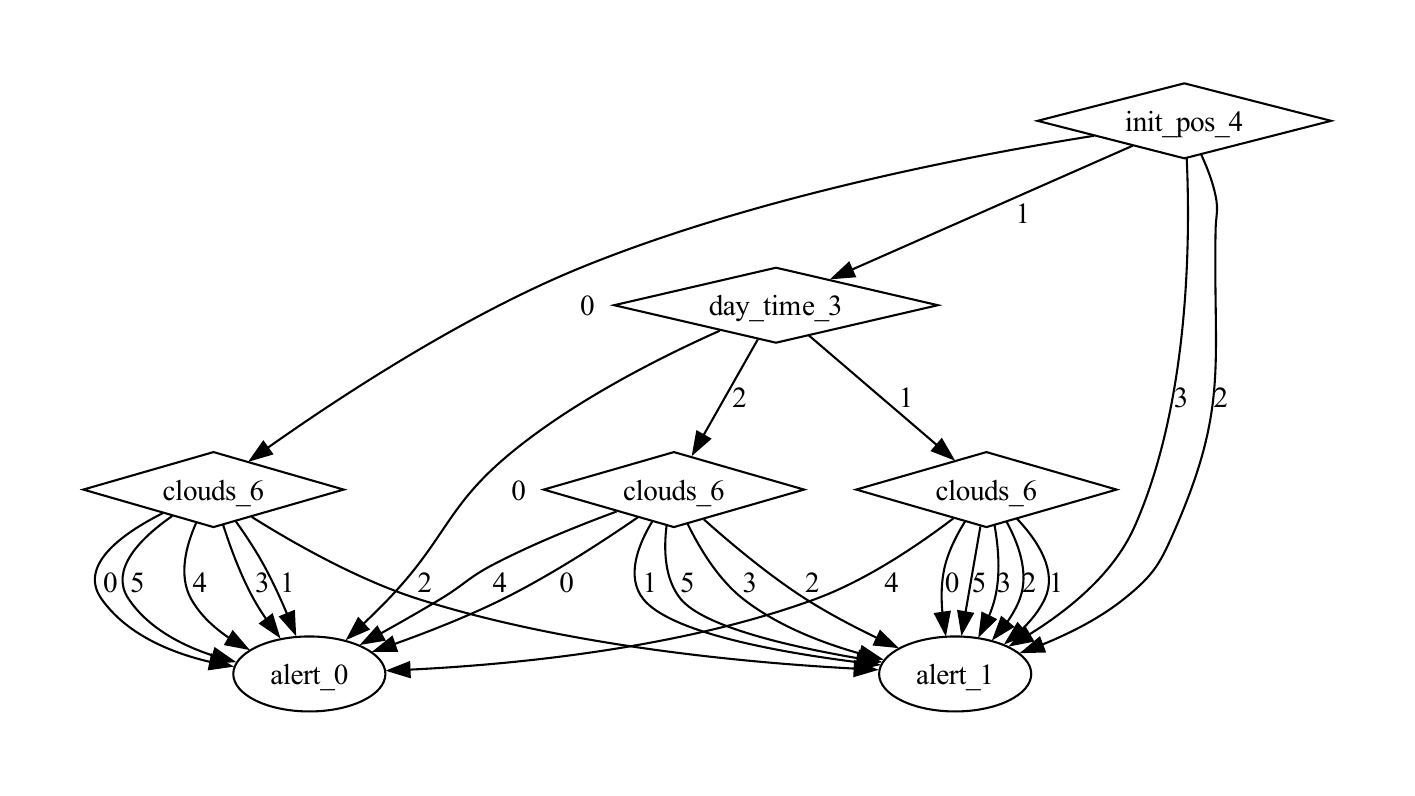}
  \captionof{figure}{Pareto-optimal decision tree with highest accuracy measure in the AutoTaxi benchmark.}
  \label{fig:po_syn_dd}
\end{figure}

\subsubsection{AutoTaxi} This represents the benchmark AutoTaxi. This is the same as presented in the main paper. We present the feature names, weights and partition in Table \ref{tab:AutoTaxi}.

Figure~\ref{fig:AutoTaxi_tmp} plots the results.
        \begin{table}[t]
            \centering
            \begin{tabular}{|c|c|c|}
            \hline
            \textbf{Feature Names} & \textbf{Weights} &  \textbf{Output Branches}\\
            \hline

            clouds\_6 & 1 & 6\\
            day\_time\_3 & 4 & 3\\
            init\_pos\_4 & 3 & 4\\

            \hline
            \end{tabular}
            \caption{List of feature names, weights and output branches}

            \label{tab:AutoTaxi}

        \end{table}

        \begin{figure}
        \centering
        \input{summary_appendix/tikz_plot_AutoTaxi.tex} 
        
        \caption{AutoTaxi}

        \label{fig:AutoTaxi_tmp}

        \end{figure}

\subsubsection{balance scale} This represents the benchmark balance scale. This is the same as presented in the main paper. We present the feature names, weights and partition in Table \ref{tab:balance_scale}.

Figure~\ref{fig:balance_scale_tmp} plots the results.
        \begin{table}[t]
            \centering
            \begin{tabular}{|c|c|c|}
            \hline
            \textbf{Feature Names} & \textbf{Weights} &  \textbf{Output Branches}\\
            \hline

            left\_distance & 3 & 3\\
            left\_weight & 3 & 3\\
            right\_distance & 3 & 3\\
            right\_weight & 3 & 3\\

            \hline
            \end{tabular}
            \caption{List of feature names, weights and output branches}

            \label{tab:balance_scale}

        \end{table}

        \begin{figure}
        \centering
        \input{summary_appendix/tikz_plot_balance_scale.tex} 
        
        \caption{balance scale}

        \label{fig:balance_scale_tmp}

        \end{figure}

\subsubsection{car evaluation} This represents the benchmark car evaluation. This is same as presented in the main paper. We present the feature names, weights and partition in Table \ref{tab:car_evaluation}.

Figure~\ref{fig:car_evaluation_tmp} plots the results.
        \begin{table}[t]
            \centering
            \begin{tabular}{|c|c|c|}
            \hline
            \textbf{Feature Names} & \textbf{Weights} &  \textbf{Output Branches}\\
            \hline

            buying & 3 & 4\\
            doors & 3 & 4\\
            lug\_boot & 3 & 3\\
            maint & 3 & 4\\
            persons & 3 & 3\\
            safety & 3 & 3\\

            \hline
            \end{tabular}
            \caption{List of feature names, weights and output branches}

            \label{tab:car_evaluation}

        \end{table}

        \begin{figure}
        \centering
        \input{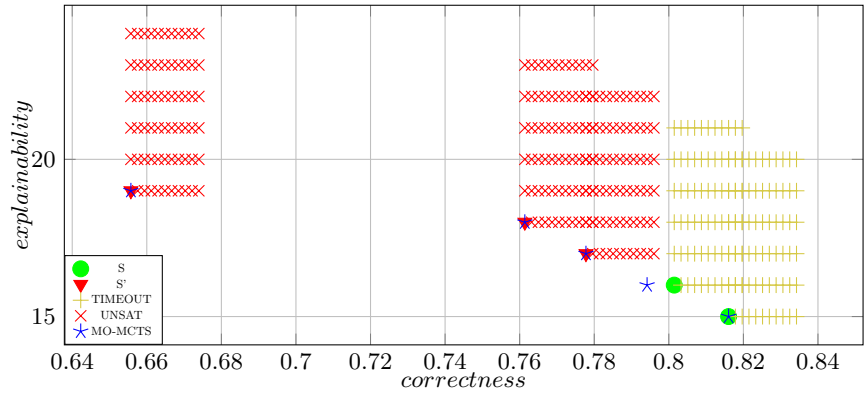} 
        
        \caption{car evaluation}

        \label{fig:car_evaluation_tmp}

        \end{figure}

\subsubsection{ecoli} This represents the benchmark ecoli. This benchmark uses a neural network trained on the dataset~\cite{ecoli} from the UCI repository. It restricts the space of decision trees to those with size at most 5. We present the feature names, weights and partition in Table \ref{tab:ecoli}.

Figure~\ref{fig:ecoli_tmp} plots the results.
        \begin{table}[t]
            \centering
            \begin{tabular}{|c|c|c|}
            \hline
            \textbf{Feature Names} & \textbf{Weights} &  \textbf{Output Branches}\\
            \hline

            aac & 3 & 3\\
            alm1 & 3 & 3\\
            alm2 & 3 & 3\\
            chg & 3 & 3\\
            gvh & 3 & 3\\
            lip & 3 & 3\\
            mcg & 3 & 3\\

            \hline
            \end{tabular}
            \caption{List of feature names, weights and output branches}

            \label{tab:ecoli}

        \end{table}

        \begin{figure}
        \centering
        \input{summary_appendix/tikz_plot_ecoli.tex} 
        
        \caption{ecoli}

        \label{fig:ecoli_tmp}

        \end{figure}

\subsubsection{credit approval} This represents the benchmark credit approval. This benchmark uses a neural network trained on the dataset~\cite{credit_approval} from the UCI repository. It restricts the space of decision trees to those with size at most 5. We present the feature names, weights and partition in Table \ref{tab:credit_approval}.

\begin{figure}
        \centering
                \begin{tikzpicture}
        \begin{axis}[
            xlabel={$correctness$},
            xlabel style={
                yshift=8pt 
            },
            ylabel={$explainability$},
            ylabel style={
                yshift=-19pt 
            },
            grid=both,
            width=\textwidth,
            height=0.5\textwidth,
            enlargelimits=0.1,
            legend style={nodes={scale=0.5, transform shape},at={(0,0)}, anchor=south west},
        ]
        
        \addplot[
                only marks,
                mark=triangle*,
                mark options={rotate=180},
                mark size=3pt,
                color=red
            ]
            coordinates {
            
            (0.8304498269896193,19)
            (0.8512110726643599,18)
            (0.8754325259515571,16)
        };
        \addlegendentry{S'}

        \addplot[
                only marks,
                mark=x,
                mark size=3pt,
                color=red
            ]
            coordinates {
            
            (0.8304498269896193,20)
            (0.8304498269896193,21)
            (0.8304498269896193,22)
            (0.8304498269896193,23)
            (0.8304498269896193,24)
            (0.8321799307958477,19)
            (0.8321799307958477,20)
            (0.8321799307958477,21)
            (0.8321799307958477,22)
            (0.8321799307958477,23)
            (0.8321799307958477,24)
            (0.8339100346020761,19)
            (0.8339100346020761,20)
            (0.8339100346020761,21)
            (0.8339100346020761,22)
            (0.8339100346020761,23)
            (0.8339100346020761,24)
            (0.8356401384083045,19)
            (0.8356401384083045,20)
            (0.8356401384083045,21)
            (0.8356401384083045,22)
            (0.8356401384083045,23)
            (0.8356401384083045,24)
            (0.8373702422145328,19)
            (0.8373702422145328,20)
            (0.8373702422145328,21)
            (0.8373702422145328,22)
            (0.8373702422145328,23)
            (0.8373702422145328,24)
            (0.8391003460207612,19)
            (0.8391003460207612,20)
            (0.8391003460207612,21)
            (0.8391003460207612,22)
            (0.8391003460207612,23)
            (0.8391003460207612,24)
            (0.8408304498269896,19)
            (0.8408304498269896,20)
            (0.8408304498269896,21)
            (0.8408304498269896,22)
            (0.8408304498269896,23)
            (0.8408304498269896,24)
            (0.842560553633218,19)
            (0.842560553633218,20)
            (0.842560553633218,21)
            (0.842560553633218,22)
            (0.842560553633218,23)
            (0.842560553633218,24)
            (0.8442906574394463,19)
            (0.8442906574394463,20)
            (0.8442906574394463,21)
            (0.8442906574394463,22)
            (0.8442906574394463,23)
            (0.8442906574394463,24)
            (0.8460207612456747,19)
            (0.8460207612456747,20)
            (0.8460207612456747,21)
            (0.8460207612456747,22)
            (0.8460207612456747,23)
            (0.8460207612456747,24)
            (0.8477508650519031,19)
            (0.8477508650519031,20)
            (0.8477508650519031,21)
            (0.8477508650519031,22)
            (0.8477508650519031,23)
            (0.8477508650519031,24)
            (0.8512110726643599,19)
            (0.8512110726643599,20)
            (0.8512110726643599,21)
            (0.8512110726643599,22)
            (0.8512110726643599,23)
            (0.8529411764705882,18)
            (0.8529411764705882,19)
            (0.8529411764705882,20)
            (0.8529411764705882,21)
            (0.8529411764705882,22)
            (0.8529411764705882,23)
            (0.8546712802768166,18)
            (0.8546712802768166,19)
            (0.8546712802768166,20)
            (0.8546712802768166,21)
            (0.8546712802768166,22)
            (0.8546712802768166,23)
            (0.856401384083045,18)
            (0.856401384083045,19)
            (0.856401384083045,20)
            (0.856401384083045,21)
            (0.856401384083045,22)
            (0.856401384083045,23)
            (0.8581314878892734,18)
            (0.8581314878892734,19)
            (0.8581314878892734,20)
            (0.8581314878892734,21)
            (0.8581314878892734,22)
            (0.8581314878892734,23)
            (0.8598615916955017,18)
            (0.8598615916955017,19)
            (0.8598615916955017,20)
            (0.8598615916955017,21)
            (0.8598615916955017,22)
            (0.8598615916955017,23)
            (0.8615916955017301,18)
            (0.8615916955017301,19)
            (0.8615916955017301,20)
            (0.8615916955017301,21)
            (0.8615916955017301,22)
            (0.8615916955017301,23)
            (0.8633217993079585,18)
            (0.8633217993079585,19)
            (0.8633217993079585,20)
            (0.8633217993079585,21)
            (0.8633217993079585,22)
            (0.8633217993079585,23)
            (0.8650519031141869,18)
            (0.8650519031141869,19)
            (0.8650519031141869,20)
            (0.8650519031141869,21)
            (0.8650519031141869,22)
            (0.8650519031141869,23)
            (0.8667820069204152,18)
            (0.8667820069204152,19)
            (0.8667820069204152,20)
            (0.8667820069204152,21)
            (0.8667820069204152,22)
            (0.8667820069204152,23)
            (0.8685121107266436,18)
            (0.8685121107266436,19)
            (0.8685121107266436,20)
            (0.8685121107266436,21)
            (0.8685121107266436,22)
            (0.8685121107266436,23)
            (0.8754325259515571,17)
            (0.8754325259515571,18)
            (0.8754325259515571,19)
            (0.8754325259515571,20)
            (0.8754325259515571,21)
            (0.8771626297577855,16)
            (0.8771626297577855,17)
            (0.8771626297577855,18)
            (0.8771626297577855,19)
            (0.8771626297577855,20)
            (0.8771626297577855,21)
            (0.8788927335640139,16)
            (0.8788927335640139,17)
            (0.8788927335640139,18)
            (0.8788927335640139,19)
            (0.8788927335640139,20)
            (0.8788927335640139,21)
            (0.8806228373702422,16)
            (0.8806228373702422,17)
            (0.8806228373702422,18)
            (0.8806228373702422,19)
            (0.8806228373702422,20)
            (0.8806228373702422,21)
            (0.8823529411764706,16)
            (0.8823529411764706,17)
            (0.8823529411764706,18)
            (0.8823529411764706,19)
            (0.8823529411764706,20)
            (0.8823529411764706,21)
            (0.884083044982699,16)
            (0.884083044982699,17)
            (0.884083044982699,18)
            (0.884083044982699,19)
            (0.884083044982699,20)
            (0.884083044982699,21)
            (0.8858131487889274,16)
            (0.8858131487889274,17)
            (0.8858131487889274,18)
            (0.8858131487889274,19)
            (0.8858131487889274,20)
            (0.8858131487889274,21)
            (0.8875432525951558,16)
            (0.8875432525951558,17)
            (0.8875432525951558,18)
            (0.8875432525951558,19)
            (0.8875432525951558,20)
            (0.8875432525951558,21)
            (0.889273356401384,16)
            (0.889273356401384,17)
            (0.889273356401384,18)
            (0.889273356401384,19)
            (0.889273356401384,20)
            (0.889273356401384,21)
            (0.8910034602076125,16)
            (0.8910034602076125,17)
            (0.8910034602076125,18)
            (0.8910034602076125,19)
            (0.8910034602076125,20)
            (0.8910034602076125,21)
            (0.8927335640138409,16)
            (0.8927335640138409,17)
            (0.8927335640138409,18)
            (0.8927335640138409,19)
            (0.8927335640138409,20)
            (0.8927335640138409,21)
        };
        \addlegendentry{UNSAT}

        \addplot[
                only marks,
                mark=star,
                mark size=3pt,
                color=blue    
            ]
            coordinates {
            
            (0.8304498269896193,19)
            (0.8512110726643599,18)
            (0.8754325259515571,15)
        };
        \addlegendentry{MO-MCTS}

        \addplot[
                only marks,
                mark=o,
                mark size=5pt,
                color=black
            ]
            coordinates {
            
            (0.8304498269896193,19)
            (0.8512110726643599,18)
            (0.8633217993079585,17)
            (0.8754325259515571,16)
        };
        \addlegendentry{Synplicate}

\end{axis}
\end{tikzpicture} 
        
        \caption{credit approval}

        \label{fig:credit_approval_tmp}

        \end{figure}
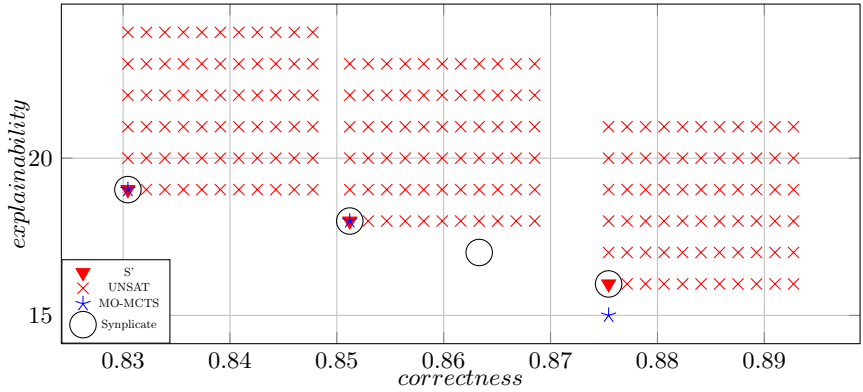

Figure~\ref{fig:credit_approval_tmp} plots the results.
        \begin{table}[t]
            \centering
            \begin{tabular}{|c|c|c|}
            \hline
            \textbf{Feature Names} & \textbf{Weights} &  \textbf{Output Branches}\\
            \hline

            a10 & 3 & 2\\
            a11 & 3 & 3\\
            a12 & 3 & 2\\
            a13 & 3 & 3\\
            a15 & 3 & 3\\
            a3 & 3 & 3\\
            a8 & 3 & 3\\
            a9 & 3 & 2\\

            \hline
            \end{tabular}
            \caption{List of feature names, weights and output branches}

            \label{tab:credit_approval}

        \end{table}

\subsubsection{yeast} This represents the benchmark yeast. This benchmark uses a neural network trained on the dataset~\cite{yeast_110} from the UCI repository. It restricts the space of decision trees to those with size at most 5. We present the feature names, weights and partition in Table \ref{tab:yeast}.

Figure~\ref{fig:yeast_tmp} plots the results.
        \begin{table}[t]
            \centering
            \begin{tabular}{|c|c|c|}
            \hline
            \textbf{Feature Names} & \textbf{Weights} &  \textbf{Output Branches}\\
            \hline

            alm & 3 & 3\\
            erl & 3 & 3\\
            gvh & 3 & 3\\
            mcg & 3 & 3\\
            mit & 3 & 3\\
            nuc & 3 & 3\\
            pox & 3 & 3\\
            vac & 3 & 3\\

            \hline
            \end{tabular}
            \caption{List of feature names, weights and output branches}

            \label{tab:yeast}

        \end{table}

        \begin{figure}
        \centering
        \input{summary_appendix/tikz_plot_yeast.tex} 
        
        \caption{yeast}

        \label{fig:yeast_tmp}

        \end{figure}

\subsubsection{tic tac toe endgame} This represents the benchmark tic tac toe endgame. This benchmark uses a neural network trained on the dataset~\cite{tic-tac-toe_endgame} from the UCI repository. It restricts the space of decision trees to those with size at most 5. We present the feature names, weights and partition in Table \ref{tab:tic_tac_toe_endgame}.

\begin{figure}
        \centering
        \input{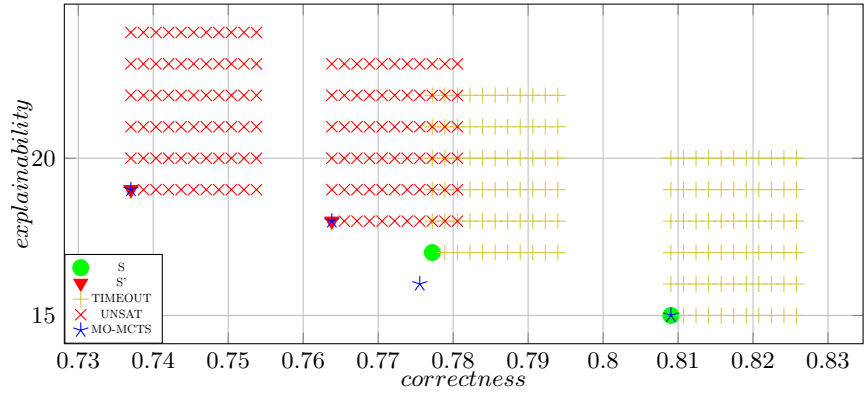} 
        
        \caption{tic tac toe endgame}

        \label{fig:tic_tac_toe_endgame_tmp}

        \end{figure}

Figure~\ref{fig:tic_tac_toe_endgame_tmp} plots the results.
        \begin{table}[t]
            \centering
            \begin{tabular}{|c|c|c|}
            \hline
            \textbf{Feature Names} & \textbf{Weights} &  \textbf{Output Branches}\\
            \hline

            bottom\_left\_square & 3 & 3\\
            bottom\_middle\_square & 3 & 3\\
            bottom\_right\_square & 3 & 3\\
            middle\_left\_square & 3 & 3\\
            middle\_middle\_square & 3 & 3\\
            middle\_right\_square & 3 & 3\\
            top\_left\_square & 3 & 3\\
            top\_middle\_square & 3 & 3\\
            top\_right\_square & 3 & 3\\

            \hline
            \end{tabular}
            \caption{List of feature names, weights and output branches}

            \label{tab:tic_tac_toe_endgame}

        \end{table}

\subsubsection{heart disease} This represents the benchmark heart disease. This benchmark uses a neural network trained on the dataset~\cite{heart_disease} from the UCI repository. It restricts the space of decision trees to those with size at most 5. We present the feature names, weights and partition in Table \ref{tab:heart_disease}.

\begin{figure}
        \centering
                \begin{tikzpicture}
        \begin{axis}[
            xlabel={$correctness$},
            xlabel style={
                yshift=8pt 
            },
            ylabel={$explainability$},
            ylabel style={
                yshift=-19pt 
            },
            grid=both,
            width=\textwidth,
            height=0.5\textwidth,
            enlargelimits=0.1,
            legend style={nodes={scale=0.5, transform shape},at={(0,0)}, anchor=south west},
        ]
        
        \addplot[
                only marks,
                mark=*,
                mark size=3pt,
                color=green
            ]
            coordinates {
            
            (0.8859375,15)
        };
        \addlegendentry{S}

        \addplot[
                only marks,
                mark=triangle*,
                mark options={rotate=180},
                mark size=3pt,
                color=red
            ]
            coordinates {
            
            (0.75625,19)
            (0.796875,18)
        };
        \addlegendentry{S'}

        \addplot[
                only marks,
                mark=+,
                mark options={rotate=180},
                mark size=3pt,
                color=yellow!80!black
            ]
            coordinates {
            
            (0.8859375,16)
            (0.8859375,17)
            (0.8859375,18)
            (0.8859375,19)
            (0.8859375,20)
            (0.8875,15)
            (0.8875,16)
            (0.8875,17)
            (0.8875,18)
            (0.8875,19)
            (0.8875,20)
            (0.8890625,15)
            (0.8890625,16)
            (0.8890625,17)
            (0.8890625,18)
            (0.8890625,19)
            (0.8890625,20)
            (0.890625,15)
            (0.890625,16)
            (0.890625,17)
            (0.890625,18)
            (0.890625,19)
            (0.890625,20)
            (0.8921875,15)
            (0.8921875,16)
            (0.8921875,17)
            (0.8921875,18)
            (0.8921875,19)
            (0.8921875,20)
            (0.89375,15)
            (0.89375,16)
            (0.89375,17)
            (0.89375,18)
            (0.89375,19)
            (0.89375,20)
            (0.8953125,15)
            (0.8953125,16)
            (0.8953125,17)
            (0.8953125,18)
            (0.8953125,19)
            (0.8953125,20)
            (0.896875,15)
            (0.896875,16)
            (0.896875,17)
            (0.896875,18)
            (0.896875,19)
            (0.896875,20)
            (0.8984375,15)
            (0.8984375,16)
            (0.8984375,17)
            (0.8984375,18)
            (0.8984375,19)
            (0.8984375,20)
            (0.9,15)
            (0.9,16)
            (0.9,17)
            (0.9,18)
            (0.9,19)
            (0.9,20)
            (0.9015625,15)
            (0.9015625,16)
            (0.9015625,17)
            (0.9015625,18)
            (0.9015625,19)
            (0.9015625,20)
        };
        \addlegendentry{TIMEOUT}

        \addplot[
                only marks,
                mark=x,
                mark size=3pt,
                color=red
            ]
            coordinates {
            
            (0.75625,20)
            (0.75625,21)
            (0.75625,22)
            (0.75625,23)
            (0.75625,24)
            (0.7578125,19)
            (0.7578125,20)
            (0.7578125,21)
            (0.7578125,22)
            (0.7578125,23)
            (0.7578125,24)
            (0.759375,19)
            (0.759375,20)
            (0.759375,21)
            (0.759375,22)
            (0.759375,23)
            (0.759375,24)
            (0.7609375,19)
            (0.7609375,20)
            (0.7609375,21)
            (0.7609375,22)
            (0.7609375,23)
            (0.7609375,24)
            (0.7625,19)
            (0.7625,20)
            (0.7625,21)
            (0.7625,22)
            (0.7625,23)
            (0.7625,24)
            (0.7640625,19)
            (0.7640625,20)
            (0.7640625,21)
            (0.7640625,22)
            (0.7640625,23)
            (0.7640625,24)
            (0.765625,19)
            (0.765625,20)
            (0.765625,21)
            (0.765625,22)
            (0.765625,23)
            (0.765625,24)
            (0.7671875,19)
            (0.7671875,20)
            (0.7671875,21)
            (0.7671875,22)
            (0.7671875,23)
            (0.7671875,24)
            (0.76875,19)
            (0.76875,20)
            (0.76875,21)
            (0.76875,22)
            (0.76875,23)
            (0.76875,24)
            (0.7703125,19)
            (0.7703125,20)
            (0.7703125,21)
            (0.7703125,22)
            (0.7703125,23)
            (0.7703125,24)
            (0.771875,19)
            (0.771875,20)
            (0.771875,21)
            (0.771875,22)
            (0.771875,23)
            (0.771875,24)
            (0.796875,19)
            (0.796875,20)
            (0.796875,21)
            (0.796875,22)
            (0.796875,23)
            (0.7984375,18)
            (0.7984375,19)
            (0.7984375,20)
            (0.7984375,21)
            (0.7984375,22)
            (0.7984375,23)
            (0.8,18)
            (0.8,19)
            (0.8,20)
            (0.8,21)
            (0.8,22)
            (0.8,23)
            (0.8015625,18)
            (0.8015625,19)
            (0.8015625,20)
            (0.8015625,21)
            (0.8015625,22)
            (0.8015625,23)
            (0.803125,18)
            (0.803125,19)
            (0.803125,20)
            (0.803125,21)
            (0.803125,22)
            (0.803125,23)
            (0.8046875,18)
            (0.8046875,19)
            (0.8046875,20)
            (0.8046875,21)
            (0.8046875,22)
            (0.8046875,23)
            (0.80625,18)
            (0.80625,19)
            (0.80625,20)
            (0.80625,21)
            (0.80625,22)
            (0.80625,23)
            (0.8078125,18)
            (0.8078125,19)
            (0.8078125,20)
            (0.8078125,21)
            (0.8078125,22)
            (0.8078125,23)
            (0.809375,18)
            (0.809375,19)
            (0.809375,20)
            (0.809375,21)
            (0.809375,22)
            (0.809375,23)
            (0.8109375,18)
            (0.8109375,19)
            (0.8109375,20)
            (0.8109375,21)
            (0.8109375,22)
            (0.8109375,23)
            (0.8125,18)
            (0.8125,19)
            (0.8125,20)
            (0.8125,21)
            (0.8125,22)
            (0.8125,23)
        };
        \addlegendentry{UNSAT}

        \addplot[
                only marks,
                mark=star,
                mark size=3pt,
                color=blue    
            ]
            coordinates {
            
            (0.75625,19)
            (0.7828125,18)
            (0.86875,15)
        };
        \addlegendentry{MO-MCTS}

        \addplot[
                only marks,
                mark=o,
                mark size=5pt,
                color=black
            ]
            coordinates {
            
            (0.75625,19)
            (0.796875,18)
            (0.8453125,17)
            (0.878125,16)
            (0.9,15)
        };
        \addlegendentry{Synplicate}

\end{axis}
\end{tikzpicture} 
        
        \caption{heart disease}

        \label{fig:heart_disease_tmp}

        \end{figure}
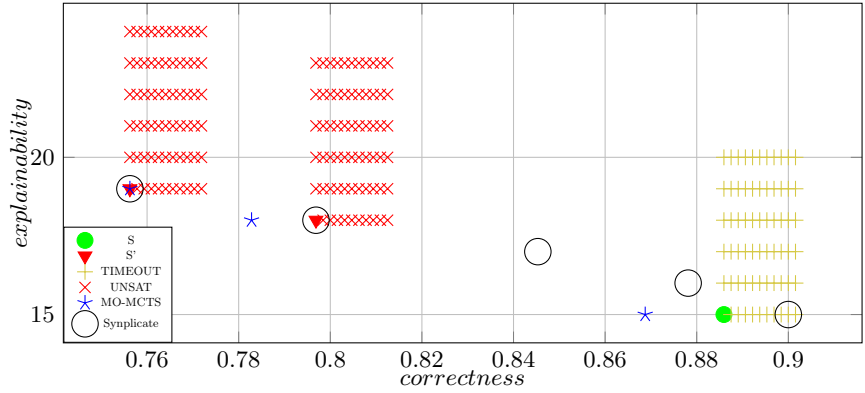

Figure~\ref{fig:heart_disease_tmp} plots the results.
        \begin{table}[t]
            \centering
            \begin{tabular}{|c|c|c|}
            \hline
            \textbf{Feature Names} & \textbf{Weights} &  \textbf{Output Branches}\\
            \hline

            age & 3 & 3\\
            chol & 3 & 3\\
            cp & 3 & 3\\
            exang & 3 & 3\\
            fbs & 3 & 3\\
            oldpeak & 3 & 3\\
            restecg & 3 & 3\\
            sex & 3 & 3\\
            slope & 3 & 3\\
            thalach & 3 & 3\\
            trestbps & 3 & 3\\

            \hline
            \end{tabular}
            \caption{List of feature names, weights and output branches}

            \label{tab:heart_disease}

        \end{table}

\subsubsection{early stage diabetes risk prediction} This represents the benchmark early stage diabetes risk prediction. This benchmark uses a neural network trained on the dataset~\cite{early_stage_diabetes_risk_prediction} from the UCI repository. It restricts the space of decision trees to those with size at most 5. We present the feature names, weights and partition in Table \ref{tab:early_stage_diabetes_risk_prediction}.

\begin{figure}
        \centering
        \input{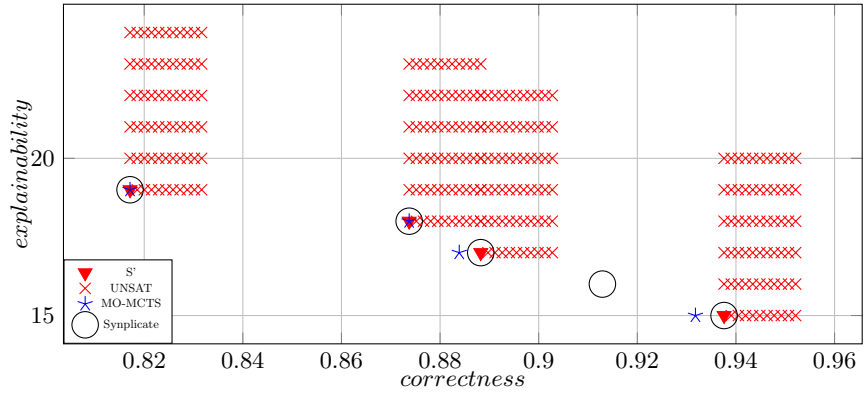} 
        
        \caption{early stage diabetes risk prediction}

        \label{fig:early_stage_diabetes_risk_prediction_tmp}

        \end{figure}

Figure~\ref{fig:early_stage_diabetes_risk_prediction_tmp} plots the results.
        \begin{table}[t]
            \centering
            \begin{tabular}{|c|c|c|}
            \hline
            \textbf{Feature Names} & \textbf{Weights} &  \textbf{Output Branches}\\
            \hline

            age & 3 & 3\\
            alopecia & 3 & 2\\
            delayed\_healing & 3 & 2\\
            gender & 3 & 2\\
            genital\_thrush & 3 & 2\\
            irritability & 3 & 2\\
            itching & 3 & 2\\
            muscle\_stiffness & 3 & 2\\
            obesity & 3 & 2\\
            partial\_paresis & 3 & 2\\
            polydipsia & 3 & 2\\
            polyphagia & 3 & 2\\
            polyuria & 3 & 2\\
            sudden\_weight\_loss & 3 & 2\\
            visual\_blurring & 3 & 2\\
            weakness & 3 & 2\\

            \hline
            \end{tabular}
            \caption{List of feature names, weights and output branches}

            \label{tab:early_stage_diabetes_risk_prediction}

        \end{table}

\subsubsection{heart failure clinical records} This represents the benchmark heart failure clinical records. This benchmark uses a neural network trained on the dataset~\cite{heart_failure_clinical_records_519} from the UCI repository. It restricts the space of decision trees to those with size at most 5. We present the feature names, weights and partition in Table \ref{tab:heart_failure_clinical_records}.

\begin{figure}
        \centering
                \begin{tikzpicture}
        \begin{axis}[
            xlabel={$correctness$},
            xlabel style={
                yshift=8pt 
            },
            ylabel={$explainability$},
            ylabel style={
                yshift=-19pt 
            },
            grid=both,
            width=\textwidth,
            height=0.5\textwidth,
            enlargelimits=0.1,
            legend style={nodes={scale=0.5, transform shape},at={(0,0)}, anchor=south west},
        ]
        
        \addplot[
                only marks,
                mark=*,
                mark size=3pt,
                color=green
            ]
            coordinates {
            
            (0.8880248833592534,15)
        };
        \addlegendentry{S}

        \addplot[
                only marks,
                mark=triangle*,
                mark options={rotate=180},
                mark size=3pt,
                color=red
            ]
            coordinates {
            
            (0.8180404354587869,19)
            (0.8460342146189735,18)
        };
        \addlegendentry{S'}

        \addplot[
                only marks,
                mark=+,
                mark options={rotate=180},
                mark size=3pt,
                color=yellow!80!black
            ]
            coordinates {
            
            (0.8880248833592534,16)
            (0.8880248833592534,17)
            (0.8880248833592534,18)
            (0.8880248833592534,19)
            (0.8880248833592534,20)
            (0.8895800933125972,15)
            (0.8895800933125972,16)
            (0.8895800933125972,17)
            (0.8895800933125972,18)
            (0.8895800933125972,19)
            (0.8895800933125972,20)
            (0.8911353032659409,15)
            (0.8911353032659409,16)
            (0.8911353032659409,17)
            (0.8911353032659409,18)
            (0.8911353032659409,19)
            (0.8911353032659409,20)
            (0.8926905132192846,15)
            (0.8926905132192846,16)
            (0.8926905132192846,17)
            (0.8926905132192846,18)
            (0.8926905132192846,19)
            (0.8926905132192846,20)
            (0.8942457231726283,15)
            (0.8942457231726283,16)
            (0.8942457231726283,17)
            (0.8942457231726283,18)
            (0.8942457231726283,19)
            (0.8942457231726283,20)
            (0.895800933125972,15)
            (0.895800933125972,16)
            (0.895800933125972,17)
            (0.895800933125972,18)
            (0.895800933125972,19)
            (0.895800933125972,20)
            (0.8973561430793157,15)
            (0.8973561430793157,16)
            (0.8973561430793157,17)
            (0.8973561430793157,18)
            (0.8973561430793157,19)
            (0.8973561430793157,20)
            (0.8989113530326595,15)
            (0.8989113530326595,16)
            (0.8989113530326595,17)
            (0.8989113530326595,18)
            (0.8989113530326595,19)
            (0.8989113530326595,20)
            (0.9004665629860031,15)
            (0.9004665629860031,16)
            (0.9004665629860031,17)
            (0.9004665629860031,18)
            (0.9004665629860031,19)
            (0.9004665629860031,20)
            (0.9020217729393468,15)
            (0.9020217729393468,16)
            (0.9020217729393468,17)
            (0.9020217729393468,18)
            (0.9020217729393468,19)
            (0.9020217729393468,20)
            (0.9035769828926905,15)
            (0.9035769828926905,16)
            (0.9035769828926905,17)
            (0.9035769828926905,18)
            (0.9035769828926905,19)
            (0.9035769828926905,20)
        };
        \addlegendentry{TIMEOUT}

        \addplot[
                only marks,
                mark=x,
                mark size=3pt,
                color=red
            ]
            coordinates {
            
            (0.8180404354587869,20)
            (0.8180404354587869,21)
            (0.8180404354587869,22)
            (0.8180404354587869,23)
            (0.8180404354587869,24)
            (0.8195956454121306,19)
            (0.8195956454121306,20)
            (0.8195956454121306,21)
            (0.8195956454121306,22)
            (0.8195956454121306,23)
            (0.8195956454121306,24)
            (0.8211508553654744,19)
            (0.8211508553654744,20)
            (0.8211508553654744,21)
            (0.8211508553654744,22)
            (0.8211508553654744,23)
            (0.8211508553654744,24)
            (0.8227060653188181,19)
            (0.8227060653188181,20)
            (0.8227060653188181,21)
            (0.8227060653188181,22)
            (0.8227060653188181,23)
            (0.8227060653188181,24)
            (0.8242612752721618,19)
            (0.8242612752721618,20)
            (0.8242612752721618,21)
            (0.8242612752721618,22)
            (0.8242612752721618,23)
            (0.8242612752721618,24)
            (0.8258164852255054,19)
            (0.8258164852255054,20)
            (0.8258164852255054,21)
            (0.8258164852255054,22)
            (0.8258164852255054,23)
            (0.8258164852255054,24)
            (0.8273716951788491,19)
            (0.8273716951788491,20)
            (0.8273716951788491,21)
            (0.8273716951788491,22)
            (0.8273716951788491,23)
            (0.8273716951788491,24)
            (0.8289269051321928,19)
            (0.8289269051321928,20)
            (0.8289269051321928,21)
            (0.8289269051321928,22)
            (0.8289269051321928,23)
            (0.8289269051321928,24)
            (0.8304821150855366,19)
            (0.8304821150855366,20)
            (0.8304821150855366,21)
            (0.8304821150855366,22)
            (0.8304821150855366,23)
            (0.8304821150855366,24)
            (0.8320373250388803,19)
            (0.8320373250388803,20)
            (0.8320373250388803,21)
            (0.8320373250388803,22)
            (0.8320373250388803,23)
            (0.8320373250388803,24)
            (0.833592534992224,19)
            (0.833592534992224,20)
            (0.833592534992224,21)
            (0.833592534992224,22)
            (0.833592534992224,23)
            (0.833592534992224,24)
            (0.8460342146189735,19)
            (0.8460342146189735,20)
            (0.8460342146189735,21)
            (0.8460342146189735,22)
            (0.8460342146189735,23)
            (0.8475894245723172,18)
            (0.8475894245723172,19)
            (0.8475894245723172,20)
            (0.8475894245723172,21)
            (0.8475894245723172,22)
            (0.8475894245723172,23)
            (0.8491446345256609,18)
            (0.8491446345256609,19)
            (0.8491446345256609,20)
            (0.8491446345256609,21)
            (0.8491446345256609,22)
            (0.8491446345256609,23)
            (0.8506998444790047,18)
            (0.8506998444790047,19)
            (0.8506998444790047,20)
            (0.8506998444790047,21)
            (0.8506998444790047,22)
            (0.8506998444790047,23)
            (0.8522550544323484,18)
            (0.8522550544323484,19)
            (0.8522550544323484,20)
            (0.8522550544323484,21)
            (0.8522550544323484,22)
            (0.8522550544323484,23)
            (0.8538102643856921,18)
            (0.8538102643856921,19)
            (0.8538102643856921,20)
            (0.8538102643856921,21)
            (0.8538102643856921,22)
            (0.8538102643856921,23)
            (0.8553654743390358,18)
            (0.8553654743390358,19)
            (0.8553654743390358,20)
            (0.8553654743390358,21)
            (0.8553654743390358,22)
            (0.8553654743390358,23)
            (0.8569206842923794,18)
            (0.8569206842923794,19)
            (0.8569206842923794,20)
            (0.8569206842923794,21)
            (0.8569206842923794,22)
            (0.8569206842923794,23)
            (0.8584758942457231,18)
            (0.8584758942457231,19)
            (0.8584758942457231,20)
            (0.8584758942457231,21)
            (0.8584758942457231,22)
            (0.8584758942457231,23)
            (0.8600311041990669,18)
            (0.8600311041990669,19)
            (0.8600311041990669,20)
            (0.8600311041990669,21)
            (0.8600311041990669,22)
            (0.8600311041990669,23)
            (0.8615863141524106,18)
            (0.8615863141524106,19)
            (0.8615863141524106,20)
            (0.8615863141524106,21)
            (0.8615863141524106,22)
            (0.8615863141524106,23)
        };
        \addlegendentry{UNSAT}

        \addplot[
                only marks,
                mark=star,
                mark size=3pt,
                color=blue    
            ]
            coordinates {
            
            (0.8180404354587869,19)
            (0.8460342146189735,18)
            (0.8880248833592534,15)
        };
        \addlegendentry{MO-MCTS}

\end{axis}
\end{tikzpicture} 
        
        \caption{heart failure clinical records}

        \label{fig:heart_failure_clinical_records_tmp}

        \end{figure}
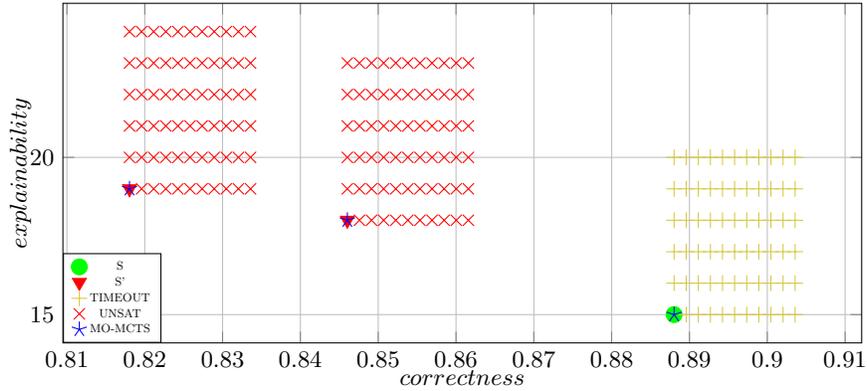

Figure~\ref{fig:heart_failure_clinical_records_tmp} plots the results.
        \begin{table}[t]
            \centering
            \begin{tabular}{|c|c|c|}
            \hline
            \textbf{Feature Names} & \textbf{Weights} &  \textbf{Output Branches}\\
            \hline

            age & 3 & 3\\
            anaemia & 3 & 3\\
            creatinine\_phosphokinase & 3 & 3\\
            diabetes & 3 & 3\\
            ejection\_fraction & 3 & 3\\
            high\_blood\_pressure & 3 & 3\\
            platelets & 3 & 3\\
            serum\_creatinine & 3 & 3\\
            serum\_sodium & 3 & 3\\
            sex & 3 & 3\\
            smoking & 3 & 3\\
            time & 3 & 3\\

            \hline
            \end{tabular}
            \caption{List of feature names, weights and output branches}

            \label{tab:heart_failure_clinical_records}

        \end{table}

\subsubsection{national poll on healthy aging  npha } This represents the benchmark national poll on healthy aging  npha . This benchmark uses a neural network trained on the dataset~\cite{national_poll_on_healthy_aging_(npha)_936} from the UCI repository. It restricts the space of decision trees to those with size at most 5. We present the feature names, weights and partition in Table \ref{tab:national_poll_on_healthy_aging__npha_}.

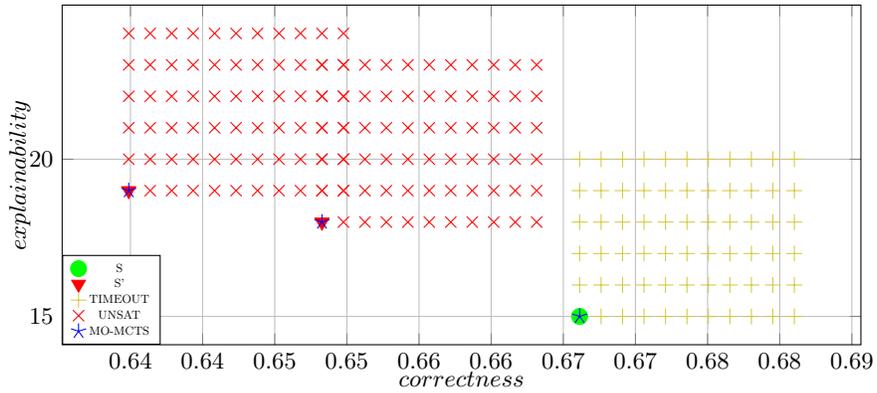
\begin{figure}
        \centering
                \begin{tikzpicture}
        \begin{axis}[
            xlabel={$correctness$},
            xlabel style={
                yshift=8pt 
            },
            ylabel={$explainability$},
            ylabel style={
                yshift=-19pt 
            },
            grid=both,
            width=\textwidth,
            height=0.5\textwidth,
            enlargelimits=0.1,
            legend style={nodes={scale=0.5, transform shape},at={(0,0)}, anchor=south west},
        ]
        
        \addplot[
                only marks,
                mark=*,
                mark size=3pt,
                color=green
            ]
            coordinates {
            
            (0.6711309523809523,15)
        };
        \addlegendentry{S}

        \addplot[
                only marks,
                mark=triangle*,
                mark options={rotate=180},
                mark size=3pt,
                color=red
            ]
            coordinates {
            
            (0.6398809523809523,19)
            (0.6532738095238095,18)
        };
        \addlegendentry{S'}

        \addplot[
                only marks,
                mark=+,
                mark options={rotate=180},
                mark size=3pt,
                color=yellow!80!black
            ]
            coordinates {
            
            (0.6711309523809523,16)
            (0.6711309523809523,17)
            (0.6711309523809523,18)
            (0.6711309523809523,19)
            (0.6711309523809523,20)
            (0.6726190476190477,15)
            (0.6726190476190477,16)
            (0.6726190476190477,17)
            (0.6726190476190477,18)
            (0.6726190476190477,19)
            (0.6726190476190477,20)
            (0.6741071428571429,15)
            (0.6741071428571429,16)
            (0.6741071428571429,17)
            (0.6741071428571429,18)
            (0.6741071428571429,19)
            (0.6741071428571429,20)
            (0.6755952380952381,15)
            (0.6755952380952381,16)
            (0.6755952380952381,17)
            (0.6755952380952381,18)
            (0.6755952380952381,19)
            (0.6755952380952381,20)
            (0.6770833333333334,15)
            (0.6770833333333334,16)
            (0.6770833333333334,17)
            (0.6770833333333334,18)
            (0.6770833333333334,19)
            (0.6770833333333334,20)
            (0.6785714285714286,15)
            (0.6785714285714286,16)
            (0.6785714285714286,17)
            (0.6785714285714286,18)
            (0.6785714285714286,19)
            (0.6785714285714286,20)
            (0.6800595238095238,15)
            (0.6800595238095238,16)
            (0.6800595238095238,17)
            (0.6800595238095238,18)
            (0.6800595238095238,19)
            (0.6800595238095238,20)
            (0.6815476190476191,15)
            (0.6815476190476191,16)
            (0.6815476190476191,17)
            (0.6815476190476191,18)
            (0.6815476190476191,19)
            (0.6815476190476191,20)
            (0.6830357142857143,15)
            (0.6830357142857143,16)
            (0.6830357142857143,17)
            (0.6830357142857143,18)
            (0.6830357142857143,19)
            (0.6830357142857143,20)
            (0.6845238095238095,15)
            (0.6845238095238095,16)
            (0.6845238095238095,17)
            (0.6845238095238095,18)
            (0.6845238095238095,19)
            (0.6845238095238095,20)
            (0.6860119047619048,15)
            (0.6860119047619048,16)
            (0.6860119047619048,17)
            (0.6860119047619048,18)
            (0.6860119047619048,19)
            (0.6860119047619048,20)
        };
        \addlegendentry{TIMEOUT}

        \addplot[
                only marks,
                mark=x,
                mark size=3pt,
                color=red
            ]
            coordinates {
            
            (0.6398809523809523,20)
            (0.6398809523809523,21)
            (0.6398809523809523,22)
            (0.6398809523809523,23)
            (0.6398809523809523,24)
            (0.6413690476190477,19)
            (0.6413690476190477,20)
            (0.6413690476190477,21)
            (0.6413690476190477,22)
            (0.6413690476190477,23)
            (0.6413690476190477,24)
            (0.6428571428571429,19)
            (0.6428571428571429,20)
            (0.6428571428571429,21)
            (0.6428571428571429,22)
            (0.6428571428571429,23)
            (0.6428571428571429,24)
            (0.6443452380952381,19)
            (0.6443452380952381,20)
            (0.6443452380952381,21)
            (0.6443452380952381,22)
            (0.6443452380952381,23)
            (0.6443452380952381,24)
            (0.6458333333333334,19)
            (0.6458333333333334,20)
            (0.6458333333333334,21)
            (0.6458333333333334,22)
            (0.6458333333333334,23)
            (0.6458333333333334,24)
            (0.6473214285714286,19)
            (0.6473214285714286,20)
            (0.6473214285714286,21)
            (0.6473214285714286,22)
            (0.6473214285714286,23)
            (0.6473214285714286,24)
            (0.6488095238095238,19)
            (0.6488095238095238,20)
            (0.6488095238095238,21)
            (0.6488095238095238,22)
            (0.6488095238095238,23)
            (0.6488095238095238,24)
            (0.6502976190476191,19)
            (0.6502976190476191,20)
            (0.6502976190476191,21)
            (0.6502976190476191,22)
            (0.6502976190476191,23)
            (0.6502976190476191,24)
            (0.6517857142857143,19)
            (0.6517857142857143,20)
            (0.6517857142857143,21)
            (0.6517857142857143,22)
            (0.6517857142857143,23)
            (0.6517857142857143,24)
            (0.6532738095238095,19)
            (0.6532738095238095,19)
            (0.6532738095238095,20)
            (0.6532738095238095,20)
            (0.6532738095238095,21)
            (0.6532738095238095,21)
            (0.6532738095238095,22)
            (0.6532738095238095,22)
            (0.6532738095238095,23)
            (0.6532738095238095,23)
            (0.6532738095238095,24)
            (0.6547619047619048,18)
            (0.6547619047619048,19)
            (0.6547619047619048,19)
            (0.6547619047619048,20)
            (0.6547619047619048,20)
            (0.6547619047619048,21)
            (0.6547619047619048,21)
            (0.6547619047619048,22)
            (0.6547619047619048,22)
            (0.6547619047619048,23)
            (0.6547619047619048,23)
            (0.6547619047619048,24)
            (0.65625,18)
            (0.65625,19)
            (0.65625,20)
            (0.65625,21)
            (0.65625,22)
            (0.65625,23)
            (0.6577380952380952,18)
            (0.6577380952380952,19)
            (0.6577380952380952,20)
            (0.6577380952380952,21)
            (0.6577380952380952,22)
            (0.6577380952380952,23)
            (0.6592261904761905,18)
            (0.6592261904761905,19)
            (0.6592261904761905,20)
            (0.6592261904761905,21)
            (0.6592261904761905,22)
            (0.6592261904761905,23)
            (0.6607142857142857,18)
            (0.6607142857142857,19)
            (0.6607142857142857,20)
            (0.6607142857142857,21)
            (0.6607142857142857,22)
            (0.6607142857142857,23)
            (0.6622023809523809,18)
            (0.6622023809523809,19)
            (0.6622023809523809,20)
            (0.6622023809523809,21)
            (0.6622023809523809,22)
            (0.6622023809523809,23)
            (0.6636904761904762,18)
            (0.6636904761904762,19)
            (0.6636904761904762,20)
            (0.6636904761904762,21)
            (0.6636904761904762,22)
            (0.6636904761904762,23)
            (0.6651785714285714,18)
            (0.6651785714285714,19)
            (0.6651785714285714,20)
            (0.6651785714285714,21)
            (0.6651785714285714,22)
            (0.6651785714285714,23)
            (0.6666666666666666,18)
            (0.6666666666666666,19)
            (0.6666666666666666,20)
            (0.6666666666666666,21)
            (0.6666666666666666,22)
            (0.6666666666666666,23)
            (0.6681547619047619,18)
            (0.6681547619047619,19)
            (0.6681547619047619,20)
            (0.6681547619047619,21)
            (0.6681547619047619,22)
            (0.6681547619047619,23)
        };
        \addlegendentry{UNSAT}

        \addplot[
                only marks,
                mark=star,
                mark size=3pt,
                color=blue    
            ]
            coordinates {
            
            (0.6398809523809523,19)
            (0.6532738095238095,18)
            (0.6711309523809523,15)
        };
        \addlegendentry{MO-MCTS}

\end{axis}
\end{tikzpicture} 
        
        \caption{national poll on healthy aging  npha }

        \label{fig:national_poll_on_healthy_aging__npha__tmp}

        \end{figure}

Figure~\ref{fig:national_poll_on_healthy_aging__npha__tmp} plots the results.
        \begin{table}[t]
            \centering
            \begin{tabular}{|c|c|c|}
            \hline
            \textbf{Feature Names} & \textbf{Weights} &  \textbf{Output Branches}\\
            \hline

            age & 3 & 3\\
            bathroom\_needs\_keeps\_patient\_from\_sleeping & 3 & 3\\
            dental\_health & 3 & 3\\
            employment & 3 & 3\\
            gender & 3 & 3\\
            medication\_keeps\_patient\_from\_sleeping & 3 & 3\\
            mental\_health & 3 & 3\\
            pain\_keeps\_patient\_from\_sleeping & 3 & 3\\
            physical\_health & 3 & 3\\
            prescription\_sleep\_medication & 3 & 3\\
            race & 3 & 3\\
            stress\_keeps\_patient\_from\_sleeping & 3 & 3\\
            trouble\_sleeping & 3 & 3\\
            uknown\_keeps\_patient\_from\_sleeping & 3 & 3\\

            \hline
            \end{tabular}
            \caption{List of feature names, weights and output branches}

            \label{tab:national_poll_on_healthy_aging__npha_}

        \end{table}

\subsubsection{data1} This represents the benchmark data1. This benchmark uses a custom made black-box model. It restricts the space of decision trees to those with size at most 5. We present the feature names, weights and partition in Table \ref{tab:data1}.

Figure~\ref{fig:data1_tmp} plots the results.
        \begin{table}[t]
            \centering
            \begin{tabular}{|c|c|c|}
            \hline
            \textbf{Feature Names} & \textbf{Weights} &  \textbf{Output Branches}\\
            \hline

            f1 & 2 & 2\\
            f2 & 2 & 2\\
            f3 & 2 & 2\\
            f4 & 2 & 2\\
            f5 & 2 & 2\\

            \hline
            \end{tabular}
            \caption{List of feature names, weights and output branches}

            \label{tab:data1}

        \end{table}

        \begin{figure}
        \centering
        \input{summary_appendix/tikz_plot_data1.tex} 
        
        \caption{data1}

        \label{fig:data1_tmp}

        \end{figure}

\subsubsection{data2} This represents the benchmark data2. This benchmark uses a custom made black-box model. It restricts the space of decision trees to those with size at most 5. We present the feature names, weights and partition in Table \ref{tab:data2}.

Figure~\ref{fig:data2_tmp} plots the results.
        \begin{table}[t]
            \centering
            \begin{tabular}{|c|c|c|}
            \hline
            \textbf{Feature Names} & \textbf{Weights} &  \textbf{Output Branches}\\
            \hline

            f1 & 2 & 2\\
            f2 & 2 & 2\\
            f3 & 2 & 2\\
            f4 & 2 & 2\\
            f5 & 2 & 2\\

            \hline
            \end{tabular}
            \caption{List of feature names, weights and output branches}

            \label{tab:data2}

        \end{table}

        \begin{figure}
        \centering
        \input{summary_appendix/tikz_plot_data2.tex} 
        
        \caption{data2}

        \label{fig:data2_tmp}

        \end{figure}

\subsubsection{data218} This represents the benchmark data218. This benchmark uses a custom made black-box model. It restricts the space of decision trees to those with size at most 3. We present the feature names, weights and partition in Table \ref{tab:data218}.

Figure~\ref{fig:data218_tmp} plots the results.
        \begin{table}[t]
            \centering
            \begin{tabular}{|c|c|c|}
            \hline
            \textbf{Feature Names} & \textbf{Weights} &  \textbf{Output Branches}\\
            \hline

            f1\_2 & 3 & 3\\
            f2\_2 & 3 & 3\\
            f3\_2 & 3 & 3\\
            f4\_2 & 3 & 3\\

            \hline
            \end{tabular}
            \caption{List of feature names, weights and output branches}

            \label{tab:data218}

        \end{table}

        \begin{figure}
        \centering
                \begin{tikzpicture}
        \begin{axis}[
            xlabel={$correctness$},
            xlabel style={
                yshift=8pt 
            },
            ylabel={$explainability$},
            ylabel style={
                yshift=-19pt 
            },
            grid=both,
            width=\textwidth,
            height=0.5\textwidth,
            enlargelimits=0.1,
            legend style={nodes={scale=0.5, transform shape},at={(0,0)}, anchor=south west},
        ]
        
        \addplot[
                only marks,
                mark=*,
                mark size=3pt,
                color=green
            ]
            coordinates {
            
            (0.1085972850678733,10)
            (0.14705882352941177,9)
        };
        \addlegendentry{S}

        \addplot[
                only marks,
                mark=triangle*,
                mark options={rotate=180},
                mark size=3pt,
                color=red
            ]
            coordinates {
            
            (0.07013574660633484,11)
        };
        \addlegendentry{S'}

        \addplot[
                only marks,
                mark=+,
                mark options={rotate=180},
                mark size=3pt,
                color=yellow!80!black
            ]
            coordinates {
            
            (0.1085972850678733,11)
            (0.1085972850678733,12)
            (0.1085972850678733,13)
            (0.1085972850678733,14)
            (0.1085972850678733,15)
            (0.11085972850678733,10)
            (0.11085972850678733,11)
            (0.11085972850678733,12)
            (0.11085972850678733,13)
            (0.11085972850678733,14)
            (0.11085972850678733,15)
            (0.11312217194570136,10)
            (0.11312217194570136,11)
            (0.11312217194570136,12)
            (0.11312217194570136,13)
            (0.11312217194570136,14)
            (0.11312217194570136,15)
            (0.11538461538461539,10)
            (0.11538461538461539,11)
            (0.11538461538461539,12)
            (0.11538461538461539,13)
            (0.11538461538461539,14)
            (0.11538461538461539,15)
            (0.11764705882352941,10)
            (0.11764705882352941,11)
            (0.11764705882352941,12)
            (0.11764705882352941,13)
            (0.11764705882352941,14)
            (0.11764705882352941,15)
            (0.11990950226244344,10)
            (0.11990950226244344,11)
            (0.11990950226244344,12)
            (0.11990950226244344,13)
            (0.11990950226244344,14)
            (0.11990950226244344,15)
            (0.12217194570135746,10)
            (0.12217194570135746,11)
            (0.12217194570135746,12)
            (0.12217194570135746,13)
            (0.12217194570135746,14)
            (0.12217194570135746,15)
            (0.1244343891402715,10)
            (0.1244343891402715,11)
            (0.1244343891402715,12)
            (0.1244343891402715,13)
            (0.1244343891402715,14)
            (0.1244343891402715,15)
            (0.12669683257918551,10)
            (0.12669683257918551,11)
            (0.12669683257918551,12)
            (0.12669683257918551,13)
            (0.12669683257918551,14)
            (0.12669683257918551,15)
            (0.12895927601809956,10)
            (0.12895927601809956,11)
            (0.12895927601809956,12)
            (0.12895927601809956,13)
            (0.12895927601809956,14)
            (0.12895927601809956,15)
            (0.13122171945701358,10)
            (0.13122171945701358,11)
            (0.13122171945701358,12)
            (0.13122171945701358,13)
            (0.13122171945701358,14)
            (0.13122171945701358,15)
            (0.14705882352941177,10)
            (0.14705882352941177,11)
            (0.14705882352941177,12)
            (0.14705882352941177,13)
            (0.14705882352941177,14)
            (0.1493212669683258,9)
            (0.1493212669683258,10)
            (0.1493212669683258,11)
            (0.1493212669683258,12)
            (0.1493212669683258,13)
            (0.1493212669683258,14)
            (0.1515837104072398,9)
            (0.1515837104072398,10)
            (0.1515837104072398,11)
            (0.1515837104072398,12)
            (0.1515837104072398,13)
            (0.1515837104072398,14)
            (0.15384615384615385,9)
            (0.15384615384615385,10)
            (0.15384615384615385,11)
            (0.15384615384615385,12)
            (0.15384615384615385,13)
            (0.15384615384615385,14)
            (0.15610859728506787,9)
            (0.15610859728506787,10)
            (0.15610859728506787,11)
            (0.15610859728506787,12)
            (0.15610859728506787,13)
            (0.15610859728506787,14)
            (0.1583710407239819,9)
            (0.1583710407239819,10)
            (0.1583710407239819,11)
            (0.1583710407239819,12)
            (0.1583710407239819,13)
            (0.1583710407239819,14)
            (0.16063348416289594,9)
            (0.16063348416289594,10)
            (0.16063348416289594,11)
            (0.16063348416289594,12)
            (0.16063348416289594,13)
            (0.16063348416289594,14)
            (0.16289592760180996,9)
            (0.16289592760180996,10)
            (0.16289592760180996,11)
            (0.16289592760180996,12)
            (0.16289592760180996,13)
            (0.16289592760180996,14)
            (0.16515837104072398,9)
            (0.16515837104072398,10)
            (0.16515837104072398,11)
            (0.16515837104072398,12)
            (0.16515837104072398,13)
            (0.16515837104072398,14)
            (0.167420814479638,9)
            (0.167420814479638,10)
            (0.167420814479638,11)
            (0.167420814479638,12)
            (0.167420814479638,13)
            (0.167420814479638,14)
            (0.16968325791855204,9)
            (0.16968325791855204,10)
            (0.16968325791855204,11)
            (0.16968325791855204,12)
            (0.16968325791855204,13)
            (0.16968325791855204,14)
        };
        \addlegendentry{TIMEOUT}

        \addplot[
                only marks,
                mark=x,
                mark size=3pt,
                color=red
            ]
            coordinates {
            
            (0.07013574660633484,12)
            (0.07013574660633484,13)
            (0.07013574660633484,14)
            (0.07013574660633484,15)
            (0.07013574660633484,16)
            (0.07239819004524888,11)
            (0.07239819004524888,12)
            (0.07239819004524888,13)
            (0.07239819004524888,14)
            (0.07239819004524888,15)
            (0.07239819004524888,16)
            (0.0746606334841629,11)
            (0.0746606334841629,12)
            (0.0746606334841629,13)
            (0.0746606334841629,14)
            (0.0746606334841629,15)
            (0.0746606334841629,16)
            (0.07692307692307693,11)
            (0.07692307692307693,12)
            (0.07692307692307693,13)
            (0.07692307692307693,14)
            (0.07692307692307693,15)
            (0.07692307692307693,16)
            (0.07918552036199095,11)
            (0.07918552036199095,12)
            (0.07918552036199095,13)
            (0.07918552036199095,14)
            (0.07918552036199095,15)
            (0.07918552036199095,16)
            (0.08144796380090498,11)
            (0.08144796380090498,12)
            (0.08144796380090498,13)
            (0.08144796380090498,14)
            (0.08144796380090498,15)
            (0.08144796380090498,16)
            (0.083710407239819,11)
            (0.083710407239819,12)
            (0.083710407239819,13)
            (0.083710407239819,14)
            (0.083710407239819,15)
            (0.083710407239819,16)
            (0.08597285067873303,11)
            (0.08597285067873303,12)
            (0.08597285067873303,13)
            (0.08597285067873303,14)
            (0.08597285067873303,15)
            (0.08597285067873303,16)
            (0.08823529411764706,11)
            (0.08823529411764706,12)
            (0.08823529411764706,13)
            (0.08823529411764706,14)
            (0.08823529411764706,15)
            (0.08823529411764706,16)
            (0.09049773755656108,11)
            (0.09049773755656108,12)
            (0.09049773755656108,13)
            (0.09049773755656108,14)
            (0.09049773755656108,15)
            (0.09049773755656108,16)
            (0.09276018099547512,11)
            (0.09276018099547512,12)
            (0.09276018099547512,13)
            (0.09276018099547512,14)
            (0.09276018099547512,15)
            (0.09276018099547512,16)
        };
        \addlegendentry{UNSAT}

        \addplot[
                only marks,
                mark=star,
                mark size=3pt,
                color=blue    
            ]
            coordinates {
            
            (0.07013574660633484,11)
            (0.1085972850678733,10)
            (0.14705882352941177,9)
        };
        \addlegendentry{MO-MCTS}

\end{axis}
\end{tikzpicture} 
        
        \caption{data218}

        \label{fig:data218_tmp}

        \end{figure}

\subsubsection{random 1} This represents the benchmark random 1. This benchmark uses a custom made black-box model. It restricts the space of decision trees to those with size at most 5. We present the feature names, weights and partition in Table \ref{tab:random_1}.

Figure~\ref{fig:random_1_tmp} plots the results.
        \begin{table}[t]
            \centering
            \begin{tabular}{|c|c|c|}
            \hline
            \textbf{Feature Names} & \textbf{Weights} &  \textbf{Output Branches}\\
            \hline

            f0 & 2 & 5\\
            f1 & 2 & 2\\
            f2 & 2 & 6\\
            f3 & 2 & 5\\
            f4 & 2 & 4\\

            \hline
            \end{tabular}
            \caption{List of feature names, weights and output branches}

            \label{tab:random_1}

        \end{table}

        \begin{figure}
        \centering
        \input{summary_appendix/tikz_plot_random_1.tex} 
        
        \caption{random 1}

        \label{fig:random_1_tmp}

        \end{figure}

\subsubsection{random 2} This represents the benchmark random 2. This benchmark uses a custom made black-box model. It restricts the space of decision trees to those with size at most 5. We present the feature names, weights and partition in Table \ref{tab:random_2}.

\begin{figure}
        \centering
                \begin{tikzpicture}
        \begin{axis}[
            xlabel={$correctness$},
            xlabel style={
                yshift=8pt 
            },
            ylabel={$explainability$},
            ylabel style={
                yshift=-19pt 
            },
            grid=both,
            width=\textwidth,
            height=0.5\textwidth,
            enlargelimits=0.1,
            legend style={nodes={scale=0.5, transform shape},at={(0,0)}, anchor=south west},
        ]
        
        \addplot[
                only marks,
                mark=triangle*,
                mark options={rotate=180},
                mark size=3pt,
                color=red
            ]
            coordinates {
            
            (0.5781544256120528,19)
            (0.7702448210922788,18)
            (0.8832391713747646,16)
        };
        \addlegendentry{S'}

        \addplot[
                only marks,
                mark=x,
                mark size=3pt,
                color=red
            ]
            coordinates {
            
            (0.5781544256120528,20)
            (0.5781544256120528,21)
            (0.5781544256120528,22)
            (0.5781544256120528,23)
            (0.5781544256120528,24)
            (0.5800376647834274,19)
            (0.5800376647834274,20)
            (0.5800376647834274,21)
            (0.5800376647834274,22)
            (0.5800376647834274,23)
            (0.5800376647834274,24)
            (0.5819209039548022,19)
            (0.5819209039548022,20)
            (0.5819209039548022,21)
            (0.5819209039548022,22)
            (0.5819209039548022,23)
            (0.5819209039548022,24)
            (0.583804143126177,19)
            (0.583804143126177,20)
            (0.583804143126177,21)
            (0.583804143126177,22)
            (0.583804143126177,23)
            (0.583804143126177,24)
            (0.5856873822975518,19)
            (0.5856873822975518,20)
            (0.5856873822975518,21)
            (0.5856873822975518,22)
            (0.5856873822975518,23)
            (0.5856873822975518,24)
            (0.5875706214689266,19)
            (0.5875706214689266,20)
            (0.5875706214689266,21)
            (0.5875706214689266,22)
            (0.5875706214689266,23)
            (0.5875706214689266,24)
            (0.5894538606403014,19)
            (0.5894538606403014,20)
            (0.5894538606403014,21)
            (0.5894538606403014,22)
            (0.5894538606403014,23)
            (0.5894538606403014,24)
            (0.591337099811676,19)
            (0.591337099811676,20)
            (0.591337099811676,21)
            (0.591337099811676,22)
            (0.591337099811676,23)
            (0.591337099811676,24)
            (0.5932203389830508,19)
            (0.5932203389830508,20)
            (0.5932203389830508,21)
            (0.5932203389830508,22)
            (0.5932203389830508,23)
            (0.5932203389830508,24)
            (0.5951035781544256,19)
            (0.5951035781544256,20)
            (0.5951035781544256,21)
            (0.5951035781544256,22)
            (0.5951035781544256,23)
            (0.5951035781544256,24)
            (0.5969868173258004,19)
            (0.5969868173258004,20)
            (0.5969868173258004,21)
            (0.5969868173258004,22)
            (0.5969868173258004,23)
            (0.5969868173258004,24)
            (0.7702448210922788,19)
            (0.7702448210922788,20)
            (0.7702448210922788,21)
            (0.7702448210922788,22)
            (0.7702448210922788,23)
            (0.7721280602636534,18)
            (0.7721280602636534,19)
            (0.7721280602636534,20)
            (0.7721280602636534,21)
            (0.7721280602636534,22)
            (0.7721280602636534,23)
            (0.7740112994350282,18)
            (0.7740112994350282,19)
            (0.7740112994350282,20)
            (0.7740112994350282,21)
            (0.7740112994350282,22)
            (0.7740112994350282,23)
            (0.775894538606403,18)
            (0.775894538606403,19)
            (0.775894538606403,20)
            (0.775894538606403,21)
            (0.775894538606403,22)
            (0.775894538606403,23)
            (0.7777777777777778,18)
            (0.7777777777777778,19)
            (0.7777777777777778,20)
            (0.7777777777777778,21)
            (0.7777777777777778,22)
            (0.7777777777777778,23)
            (0.7796610169491526,18)
            (0.7796610169491526,19)
            (0.7796610169491526,20)
            (0.7796610169491526,21)
            (0.7796610169491526,22)
            (0.7796610169491526,23)
            (0.7815442561205274,18)
            (0.7815442561205274,19)
            (0.7815442561205274,20)
            (0.7815442561205274,21)
            (0.7815442561205274,22)
            (0.7815442561205274,23)
            (0.783427495291902,18)
            (0.783427495291902,19)
            (0.783427495291902,20)
            (0.783427495291902,21)
            (0.783427495291902,22)
            (0.783427495291902,23)
            (0.7853107344632768,18)
            (0.7853107344632768,19)
            (0.7853107344632768,20)
            (0.7853107344632768,21)
            (0.7853107344632768,22)
            (0.7853107344632768,23)
            (0.7871939736346516,18)
            (0.7871939736346516,19)
            (0.7871939736346516,20)
            (0.7871939736346516,21)
            (0.7871939736346516,22)
            (0.7871939736346516,23)
            (0.7890772128060264,18)
            (0.7890772128060264,19)
            (0.7890772128060264,20)
            (0.7890772128060264,21)
            (0.7890772128060264,22)
            (0.7890772128060264,23)
            (0.8832391713747646,17)
            (0.8832391713747646,18)
            (0.8832391713747646,19)
            (0.8832391713747646,20)
            (0.8832391713747646,21)
            (0.8851224105461394,16)
            (0.8851224105461394,17)
            (0.8851224105461394,18)
            (0.8851224105461394,19)
            (0.8851224105461394,20)
            (0.8851224105461394,21)
            (0.8870056497175142,16)
            (0.8870056497175142,17)
            (0.8870056497175142,18)
            (0.8870056497175142,19)
            (0.8870056497175142,20)
            (0.8870056497175142,21)
            (0.8888888888888888,16)
            (0.8888888888888888,17)
            (0.8888888888888888,18)
            (0.8888888888888888,19)
            (0.8888888888888888,20)
            (0.8888888888888888,21)
            (0.8907721280602636,16)
            (0.8907721280602636,17)
            (0.8907721280602636,18)
            (0.8907721280602636,19)
            (0.8907721280602636,20)
            (0.8907721280602636,21)
            (0.8926553672316384,16)
            (0.8926553672316384,17)
            (0.8926553672316384,18)
            (0.8926553672316384,19)
            (0.8926553672316384,20)
            (0.8926553672316384,21)
            (0.8945386064030132,16)
            (0.8945386064030132,17)
            (0.8945386064030132,18)
            (0.8945386064030132,19)
            (0.8945386064030132,20)
            (0.8945386064030132,21)
            (0.896421845574388,16)
            (0.896421845574388,17)
            (0.896421845574388,18)
            (0.896421845574388,19)
            (0.896421845574388,20)
            (0.896421845574388,21)
            (0.8983050847457628,16)
            (0.8983050847457628,17)
            (0.8983050847457628,18)
            (0.8983050847457628,19)
            (0.8983050847457628,20)
            (0.8983050847457628,21)
            (0.9001883239171374,16)
            (0.9001883239171374,17)
            (0.9001883239171374,18)
            (0.9001883239171374,19)
            (0.9001883239171374,20)
            (0.9001883239171374,21)
            (0.9020715630885122,16)
            (0.9020715630885122,17)
            (0.9020715630885122,18)
            (0.9020715630885122,19)
            (0.9020715630885122,20)
            (0.9020715630885122,21)
        };
        \addlegendentry{UNSAT}

        \addplot[
                only marks,
                mark=star,
                mark size=3pt,
                color=blue    
            ]
            coordinates {
            
            (0.5781544256120528,19)
            (0.7702448210922788,18)
            (0.8832391713747646,15)
        };
        \addlegendentry{MO-MCTS}

        \addplot[
                only marks,
                mark=o,
                mark size=5pt,
                color=black
            ]
            coordinates {
            
            (0.5781544256120528,19)
            (0.7702448210922788,18)
            (0.7909604519774012,17)
            (0.8832391713747646,16)
            (0.96045197740113,15)
        };
        \addlegendentry{Synplicate}

\end{axis}
\end{tikzpicture} 
        
        \caption{random 2}

        \label{fig:random_2_tmp}

        \end{figure}
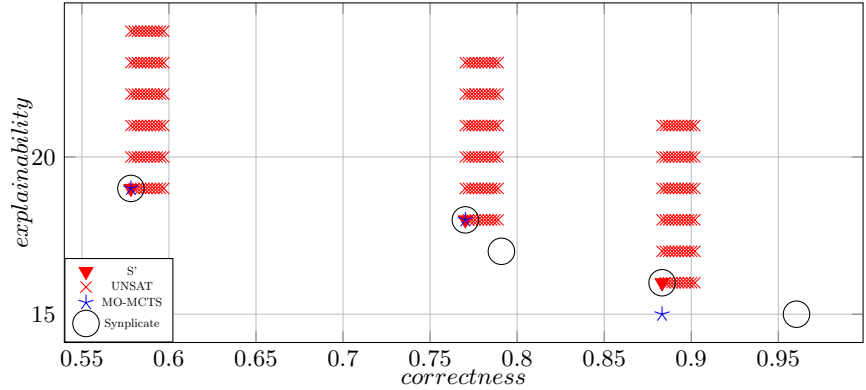

Figure~\ref{fig:random_2_tmp} plots the results.
        \begin{table}[t]
            \centering
            \begin{tabular}{|c|c|c|}
            \hline
            \textbf{Feature Names} & \textbf{Weights} &  \textbf{Output Branches}\\
            \hline

            f0 & 3 & 2\\
            f1 & 3 & 4\\
            f2 & 3 & 3\\
            f3 & 3 & 2\\
            f4 & 3 & 5\\

            \hline
            \end{tabular}
            \caption{List of feature names, weights and output branches}

            \label{tab:random_2}

        \end{table}

\subsubsection{random 7} This represents the benchmark random 7. This benchmark uses a custom made black-box model. It restricts the space of decision trees to those with size at most 3. We present the feature names, weights and partition in Table \ref{tab:random_7}.

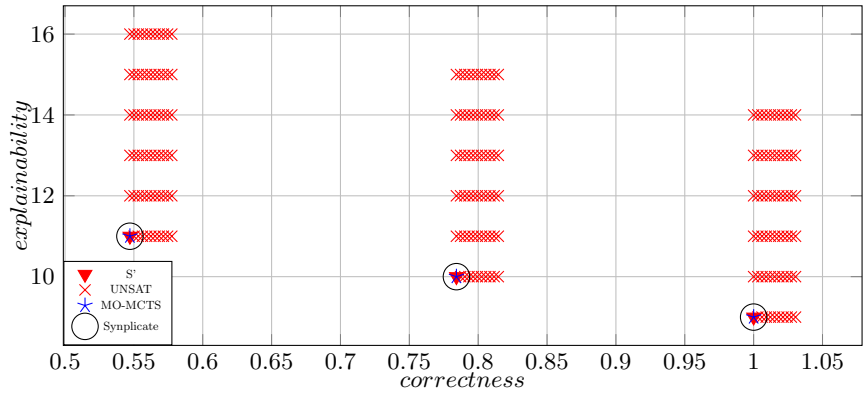
\begin{figure}
        \centering
                \begin{tikzpicture}
        \begin{axis}[
            xlabel={$correctness$},
            xlabel style={
                yshift=8pt 
            },
            ylabel={$explainability$},
            ylabel style={
                yshift=-19pt 
            },
            grid=both,
            width=\textwidth,
            height=0.5\textwidth,
            enlargelimits=0.1,
            legend style={nodes={scale=0.5, transform shape},at={(0,0)}, anchor=south west},
        ]
        
        \addplot[
                only marks,
                mark=triangle*,
                mark options={rotate=180},
                mark size=3pt,
                color=red
            ]
            coordinates {
            
            (0.547112462006079,11)
            (0.78419452887538,10)
            (1.0,9)
        };
        \addlegendentry{S'}

        \addplot[
                only marks,
                mark=x,
                mark size=3pt,
                color=red
            ]
            coordinates {
            
            (0.547112462006079,12)
            (0.547112462006079,13)
            (0.547112462006079,14)
            (0.547112462006079,15)
            (0.547112462006079,16)
            (0.5501519756838906,11)
            (0.5501519756838906,12)
            (0.5501519756838906,13)
            (0.5501519756838906,14)
            (0.5501519756838906,15)
            (0.5501519756838906,16)
            (0.5531914893617021,11)
            (0.5531914893617021,12)
            (0.5531914893617021,13)
            (0.5531914893617021,14)
            (0.5531914893617021,15)
            (0.5531914893617021,16)
            (0.5562310030395137,11)
            (0.5562310030395137,12)
            (0.5562310030395137,13)
            (0.5562310030395137,14)
            (0.5562310030395137,15)
            (0.5562310030395137,16)
            (0.5592705167173252,11)
            (0.5592705167173252,12)
            (0.5592705167173252,13)
            (0.5592705167173252,14)
            (0.5592705167173252,15)
            (0.5592705167173252,16)
            (0.5623100303951368,11)
            (0.5623100303951368,12)
            (0.5623100303951368,13)
            (0.5623100303951368,14)
            (0.5623100303951368,15)
            (0.5623100303951368,16)
            (0.5653495440729484,11)
            (0.5653495440729484,12)
            (0.5653495440729484,13)
            (0.5653495440729484,14)
            (0.5653495440729484,15)
            (0.5653495440729484,16)
            (0.5683890577507599,11)
            (0.5683890577507599,12)
            (0.5683890577507599,13)
            (0.5683890577507599,14)
            (0.5683890577507599,15)
            (0.5683890577507599,16)
            (0.5714285714285714,11)
            (0.5714285714285714,12)
            (0.5714285714285714,13)
            (0.5714285714285714,14)
            (0.5714285714285714,15)
            (0.5714285714285714,16)
            (0.574468085106383,11)
            (0.574468085106383,12)
            (0.574468085106383,13)
            (0.574468085106383,14)
            (0.574468085106383,15)
            (0.574468085106383,16)
            (0.5775075987841946,11)
            (0.5775075987841946,12)
            (0.5775075987841946,13)
            (0.5775075987841946,14)
            (0.5775075987841946,15)
            (0.5775075987841946,16)
            (0.78419452887538,11)
            (0.78419452887538,12)
            (0.78419452887538,13)
            (0.78419452887538,14)
            (0.78419452887538,15)
            (0.7872340425531915,10)
            (0.7872340425531915,11)
            (0.7872340425531915,12)
            (0.7872340425531915,13)
            (0.7872340425531915,14)
            (0.7872340425531915,15)
            (0.790273556231003,10)
            (0.790273556231003,11)
            (0.790273556231003,12)
            (0.790273556231003,13)
            (0.790273556231003,14)
            (0.790273556231003,15)
            (0.7933130699088146,10)
            (0.7933130699088146,11)
            (0.7933130699088146,12)
            (0.7933130699088146,13)
            (0.7933130699088146,14)
            (0.7933130699088146,15)
            (0.7963525835866262,10)
            (0.7963525835866262,11)
            (0.7963525835866262,12)
            (0.7963525835866262,13)
            (0.7963525835866262,14)
            (0.7963525835866262,15)
            (0.7993920972644377,10)
            (0.7993920972644377,11)
            (0.7993920972644377,12)
            (0.7993920972644377,13)
            (0.7993920972644377,14)
            (0.7993920972644377,15)
            (0.8024316109422492,10)
            (0.8024316109422492,11)
            (0.8024316109422492,12)
            (0.8024316109422492,13)
            (0.8024316109422492,14)
            (0.8024316109422492,15)
            (0.8054711246200608,10)
            (0.8054711246200608,11)
            (0.8054711246200608,12)
            (0.8054711246200608,13)
            (0.8054711246200608,14)
            (0.8054711246200608,15)
            (0.8085106382978723,10)
            (0.8085106382978723,11)
            (0.8085106382978723,12)
            (0.8085106382978723,13)
            (0.8085106382978723,14)
            (0.8085106382978723,15)
            (0.8115501519756839,10)
            (0.8115501519756839,11)
            (0.8115501519756839,12)
            (0.8115501519756839,13)
            (0.8115501519756839,14)
            (0.8115501519756839,15)
            (0.8145896656534954,10)
            (0.8145896656534954,11)
            (0.8145896656534954,12)
            (0.8145896656534954,13)
            (0.8145896656534954,14)
            (0.8145896656534954,15)
            (1.0,10)
            (1.0,11)
            (1.0,12)
            (1.0,13)
            (1.0,14)
            (1.0030395136778116,9)
            (1.0030395136778116,10)
            (1.0030395136778116,11)
            (1.0030395136778116,12)
            (1.0030395136778116,13)
            (1.0030395136778116,14)
            (1.006079027355623,9)
            (1.006079027355623,10)
            (1.006079027355623,11)
            (1.006079027355623,12)
            (1.006079027355623,13)
            (1.006079027355623,14)
            (1.0091185410334347,9)
            (1.0091185410334347,10)
            (1.0091185410334347,11)
            (1.0091185410334347,12)
            (1.0091185410334347,13)
            (1.0091185410334347,14)
            (1.012158054711246,9)
            (1.012158054711246,10)
            (1.012158054711246,11)
            (1.012158054711246,12)
            (1.012158054711246,13)
            (1.012158054711246,14)
            (1.0151975683890577,9)
            (1.0151975683890577,10)
            (1.0151975683890577,11)
            (1.0151975683890577,12)
            (1.0151975683890577,13)
            (1.0151975683890577,14)
            (1.0182370820668694,9)
            (1.0182370820668694,10)
            (1.0182370820668694,11)
            (1.0182370820668694,12)
            (1.0182370820668694,13)
            (1.0182370820668694,14)
            (1.0212765957446808,9)
            (1.0212765957446808,10)
            (1.0212765957446808,11)
            (1.0212765957446808,12)
            (1.0212765957446808,13)
            (1.0212765957446808,14)
            (1.0243161094224924,9)
            (1.0243161094224924,10)
            (1.0243161094224924,11)
            (1.0243161094224924,12)
            (1.0243161094224924,13)
            (1.0243161094224924,14)
            (1.027355623100304,9)
            (1.027355623100304,10)
            (1.027355623100304,11)
            (1.027355623100304,12)
            (1.027355623100304,13)
            (1.027355623100304,14)
            (1.0303951367781155,9)
            (1.0303951367781155,10)
            (1.0303951367781155,11)
            (1.0303951367781155,12)
            (1.0303951367781155,13)
            (1.0303951367781155,14)
        };
        \addlegendentry{UNSAT}

        \addplot[
                only marks,
                mark=star,
                mark size=3pt,
                color=blue    
            ]
            coordinates {
            
            (0.547112462006079,11)
            (0.78419452887538,10)
            (1.0,9)
        };
        \addlegendentry{MO-MCTS}

        \addplot[
                only marks,
                mark=o,
                mark size=5pt,
                color=black
            ]
            coordinates {
            
            (0.547112462006079,11)
            (0.78419452887538,10)
            (1.0,9)
        };
        \addlegendentry{Synplicate}

\end{axis}
\end{tikzpicture} 
        
        \caption{random 7}

        \label{fig:random_7_tmp}

        \end{figure}

Figure~\ref{fig:random_7_tmp} plots the results.
        \begin{table}[t]
            \centering
            \begin{tabular}{|c|c|c|}
            \hline
            \textbf{Feature Names} & \textbf{Weights} &  \textbf{Output Branches}\\
            \hline

            f0 & 3 & 2\\
            f1 & 3 & 2\\
            f2 & 3 & 2\\

            \hline
            \end{tabular}
            \caption{List of feature names, weights and output branches}

            \label{tab:random_7}

        \end{table}

\subsubsection{random 9} This represents the benchmark random 9. This benchmark uses a custom made black-box model. It restricts the space of decision trees to those with size at most 4. We present the feature names, weights and partition in Table \ref{tab:random_9}.

\begin{figure}
        \centering
        \input{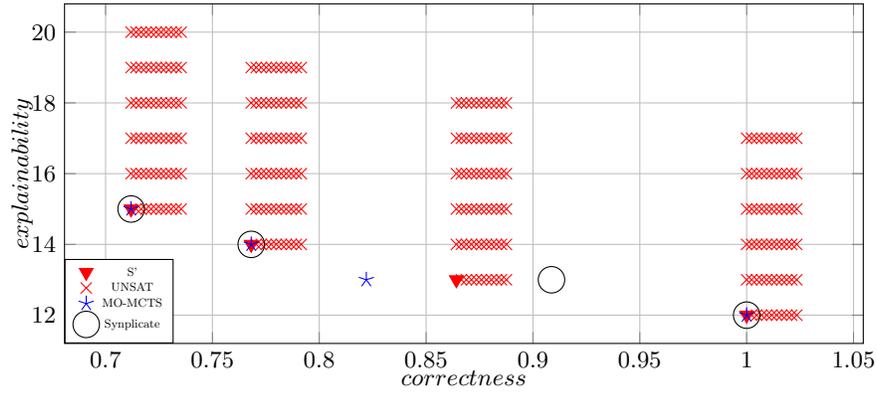} 
        
        \caption{random 9}

        \label{fig:random_9_tmp}

        \end{figure}

Figure~\ref{fig:random_9_tmp} plots the results.
        \begin{table}[t]
            \centering
            \begin{tabular}{|c|c|c|}
            \hline
            \textbf{Feature Names} & \textbf{Weights} &  \textbf{Output Branches}\\
            \hline

            f0 & 3 & 3\\
            f1 & 3 & 5\\
            f2 & 3 & 2\\
            f3 & 3 & 3\\

            \hline
            \end{tabular}
            \caption{List of feature names, weights and output branches}

            \label{tab:random_9}

        \end{table}

\subsubsection{random 10} This represents the benchmark random 10. This benchmark uses a custom made black-box model. It restricts the space of decision trees to those with size at most 5. We present the feature names, weights and partition in Table \ref{tab:random_10}.

\begin{figure}
        \centering
        \input{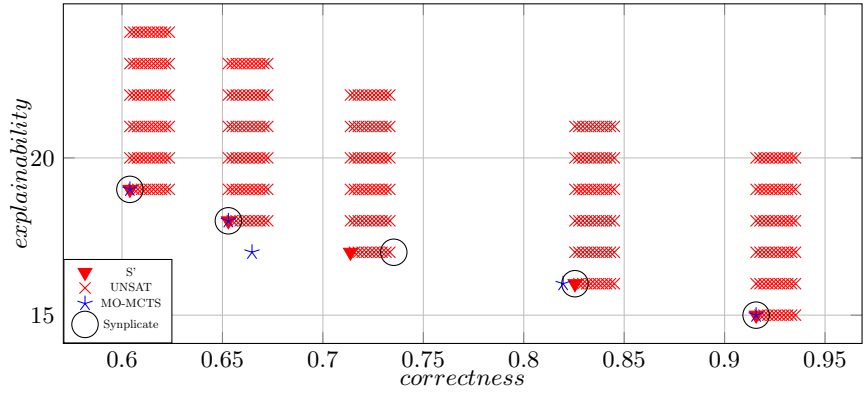} 
        
        \caption{random 10}

        \label{fig:random_10_tmp}

        \end{figure}

Figure~\ref{fig:random_10_tmp} plots the results.
        \begin{table}[t]
            \centering
            \begin{tabular}{|c|c|c|}
            \hline
            \textbf{Feature Names} & \textbf{Weights} &  \textbf{Output Branches}\\
            \hline

            f0 & 3 & 5\\
            f1 & 3 & 4\\
            f2 & 3 & 5\\
            f3 & 3 & 3\\

            \hline
            \end{tabular}
            \caption{List of feature names, weights and output branches}

            \label{tab:random_10}

        \end{table}

\section{Extended Experiments 5 mins}\label{A3}
We now present the results of running {\tool} and Synplicate with a reduced overall timeout of $5$ mins.  For {\tool}, this means we allocated 2.5 mins for the first phase (MO-MCTS), and 2.5 mins for the second phase.  Our results show that {\tool} gracefully degrades with less time being made available, but is still able to generate several LPO interpretations.  However, Synplicate practically fails to generate anything (except one PO interpretation for Balance Scale) other than for Auto Taxi.  This demonstrates the strength of {\tool} as an anytime algorithm.
\subsubsection{AutoTaxi} This represents the benchmark AutoTaxi used earlier. We restricts the space of decision trees to those with size at most 5. We present the feature names, weights and partition in Table \ref{tab:AutoTaxi}.

Figure~\ref{fig:AutoTaxi_tmp} plots the results.
        \begin{table}[t]
            \centering
            \begin{tabular}{|c|c|c|}
            \hline
            \textbf{Feature Names} & \textbf{Weights} &  \textbf{Output Branches}\\
            \hline

            clouds\_6 & 1 & 6\\
            day\_time\_3 & 4 & 3\\
            init\_pos\_4 & 3 & 4\\

            \hline
            \end{tabular}
            \caption{List of feature names, weights and output branches}

            \label{tab:AutoTaxi}

        \end{table}

        \begin{figure}
        \centering
        \input{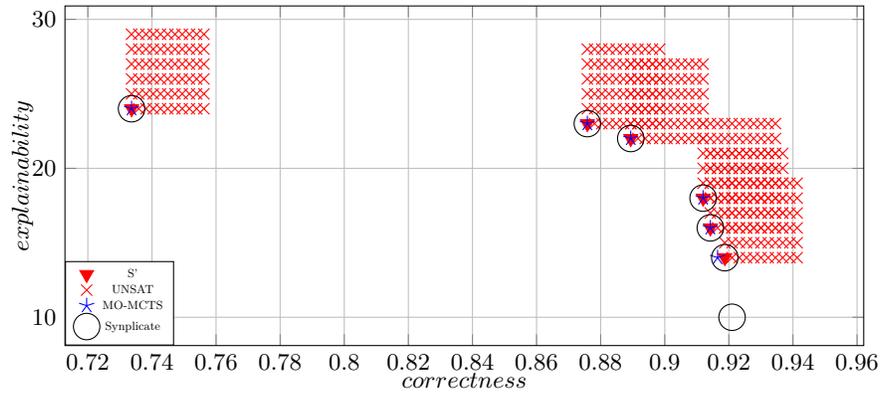} 
        
        \caption{AutoTaxi}

        \label{fig:AutoTaxi_tmp}

        \end{figure}

\subsubsection{balance scale} This represents the benchmark balance scale. This benchmark uses a neural network trained on the dataset~\cite{balance_scale} from the UCI repository. It restricts the space of decision trees to those with size at most 5. We present the feature names, weights and partition in Table \ref{tab:balance_scale}.

Figure~\ref{fig:balance_scale_tmp} plots the results.
        \begin{table}[t]
            \centering
            \begin{tabular}{|c|c|c|}
            \hline
            \textbf{Feature Names} & \textbf{Weights} &  \textbf{Output Branches}\\
            \hline

            left\_distance & 3 & 3\\
            left\_weight & 3 & 3\\
            right\_distance & 3 & 3\\
            right\_weight & 3 & 3\\

            \hline
            \end{tabular}
            \caption{List of feature names, weights and output branches}

            \label{tab:balance_scale}

        \end{table}

        \begin{figure}
        \centering
        \input{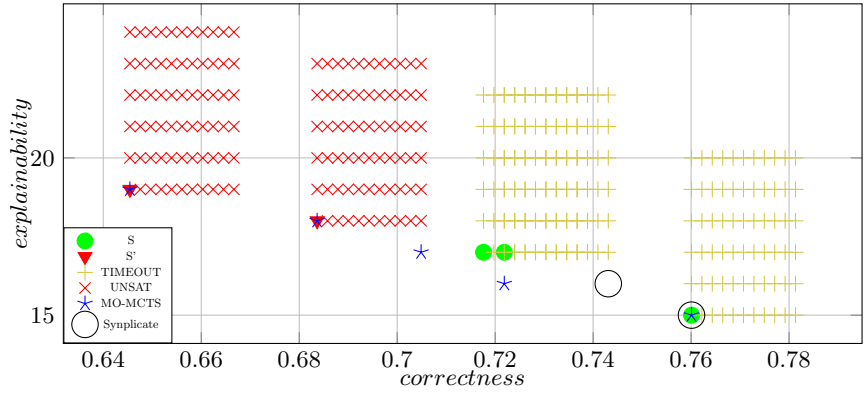} 
        
        \caption{balance scale}

        \label{fig:balance_scale_tmp}

        \end{figure}

\subsubsection{car evaluation} This represents the benchmark car evaluation. This benchmark uses a neural network trained on the dataset~\cite{car_evaluation} from the UCI repository. It restricts the space of decision trees to those with size at most 5. We present the feature names, weights and partition in Table \ref{tab:car_evaluation}.

Figure~\ref{fig:car_evaluation_tmp} plots the results.
        \begin{table}[t]
            \centering
            \begin{tabular}{|c|c|c|}
            \hline
            \textbf{Feature Names} & \textbf{Weights} &  \textbf{Output Branches}\\
            \hline

            buying & 3 & 4\\
            doors & 3 & 4\\
            lug\_boot & 3 & 3\\
            maint & 3 & 4\\
            persons & 3 & 3\\
            safety & 3 & 3\\

            \hline
            \end{tabular}
            \caption{List of feature names, weights and output branches}

            \label{tab:car_evaluation}

        \end{table}

        \begin{figure}
        \centering
        \input{summary_appendix_5mins/tikz_plot_car_evaluation.tex} 
        
        \caption{car evaluation}

        \label{fig:car_evaluation_tmp}

        \end{figure}

\subsubsection{yeast} This represents the benchmark yeast. This benchmark uses a neural network trained on the dataset~\cite{yeast_dataset} from the UCI repository. It restricts the space of decision trees to those with size at most 5. We present the feature names, weights and partition in Table \ref{tab:yeast}.

Figure~\ref{fig:yeast_tmp} plots the results.
        \begin{table}[t]
            \centering
            \begin{tabular}{|c|c|c|}
            \hline
            \textbf{Feature Names} & \textbf{Weights} &  \textbf{Output Branches}\\
            \hline

            alm & 3 & 3\\
            erl & 3 & 3\\
            gvh & 3 & 3\\
            mcg & 3 & 3\\
            mit & 3 & 3\\
            nuc & 3 & 3\\
            pox & 3 & 3\\
            vac & 3 & 3\\

            \hline
            \end{tabular}
            \caption{List of feature names, weights and output branches}

            \label{tab:yeast}

        \end{table}

        \begin{figure}
        \centering
        \input{summary_appendix_5mins/tikz_plot_yeast.tex} 
        
        \caption{yeast}

        \label{fig:yeast_tmp}

        \end{figure}

\end{document}